\documentclass[sigconf]{acmart}

\usepackage{boldline}
\usepackage{subcaption}
\usepackage{textcomp}
\usepackage{makecell}
\usepackage{multirow}
\usepackage{color}
\usepackage{soul}
\usepackage{pifont}

\AtBeginDocument{%
  }

\acmConference[KDD '25]{Proceedings of the 31st ACM SIGKDD Conference on Knowledge Discovery and Data Mining V.2}{August 3--7, 2025}{Toronto, ON, Canada}
\acmBooktitle{Proceedings of the 31st ACM SIGKDD Conference on Knowledge Discovery and Data Mining V.2 (KDD '25), August 3--7, 2025, Toronto, ON, Canada}

\begin{document}

\newcommand{\jz}[1]{{\color{red}{\bf{[JZ:]}} #1}}
\newcommand*{\red}{\textcolor{red}}
\newcommand*{\blue}{\textcolor{blue}}
\newcommand*{\orange}{\textcolor{orange}}
\newcommand*{\green}{\textcolor{green}}

\newcommand{\embmodel}{MyModel}
\newcommand{\model}{\textsc{FlexiReg}}
\newcommand{\gridGen}{GridLearner}
\newcommand{\regionGen}{AdaRegionGen}
\newcommand{\prompt}{PromptEnhancer}

\title{\model: Flexible Urban Region Representation Learning}


\author{Fengze Sun}
\affiliation{%
  \institution{School of Computing and Information System, University of Melbourne}
  \city{Melbourne}
  \country{Australia}}
\email{fengzes@student.unimelb.edu.au}

\author{Yanchuan Chang}
\affiliation{%
  \institution{School of Computing and Information System, University of Melbourne}
  \city{Melbourne}
  \country{Australia}}
\email{yanchuan.chang@unimelb.edu.au}

\author{Egemen Tanin}
\affiliation{%
  \institution{School of Computing and Information System, University of Melbourne}
  \city{Melbourne}
  \country{Australia}}
\email{etanin@.unimelb.edu.au}

\author{Shanika Karunasekera}
\affiliation{%
  \institution{School of Computing and Information System, University of Melbourne}
  \city{Melbourne}
  \country{Australia}}
\email{karus@.unimelb.edu.au}

\author{Jianzhong Qi}
\authornote{Corresponding author.}
\affiliation{%
  \institution{School of Computing and Information System, University of Melbourne}
  \city{Melbourne}
  \country{Australia}}
\email{jianzhong.qi@unimelb.edu.au}

\begin{abstract}
The increasing availability of urban data offers new opportunities for learning region representations, which can be used as input to machine learning models for downstream tasks such as check-in or crime prediction. While existing solutions have produced promising results, an issue is their fixed formation of  regions and fixed input region features, which  may not suit the needs of different downstream tasks. 
To address this limitation, we propose a  model named \model\ for urban region representation learning that is flexible with both the formation of urban regions and the input region features. \model\ is based on a spatial grid partitioning over the spatial area of interest. It learns representations for the grid cells, leveraging publicly accessible data, including POI, land use, satellite imagery, and street view imagery. We propose adaptive aggregation to fuse the cell representations and prompt learning techniques to tailor the representations towards different tasks, addressing the needs of varying formations of urban regions and downstream tasks. 
Extensive experiments on five real-world datasets demonstrate that \model\ outperforms state-of-the-art models by up to 202\% in term of the accuracy of four diverse downstream tasks using the produced urban region representations.

\end{abstract}



\maketitle

\section{Introduction}
Urban region representation learning has become increasingly popular in the community of urban computing~\cite{zhang2017urban, zheng2014urban, kdd1, su2024dualcast, su2024spatial, 10.1145/3690624.3709209, 10.1145/3696410.3714744, chang2023trajectory, chang2025k}, which aims to transform urban regions into vector representations, known as embeddings. These embeddings entail valuable insights on urban structures and properties, facilitating effective urban planning and management, such as designating functionalities for new development areas. They are also useful in various tasks related to daily life, such as crime count prediction~\cite{MGFN, HAFusion, HREP, ReCP, CGAP}.

\vspace{-1mm}
\begin{figure}[ht]
    \centering
    \begin{subfigure}[b]{\columnwidth}
        \centering
        \includegraphics[width=0.95\columnwidth]{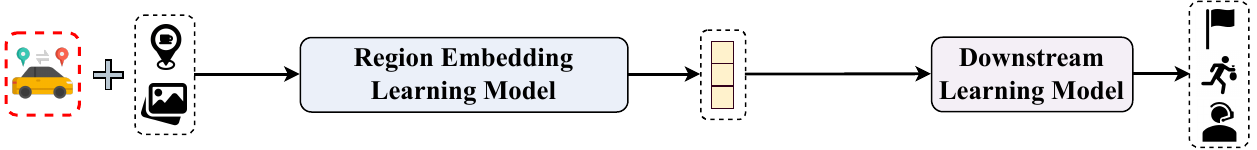}
        \vspace{-2mm}
        \caption{Representation learning for predefined region partitions}
        \label{fig:intro1}
    \end{subfigure}
    \begin{subfigure}[b]{\columnwidth}
        \vspace{2mm}
        \centering
        \includegraphics[width=0.95\columnwidth]{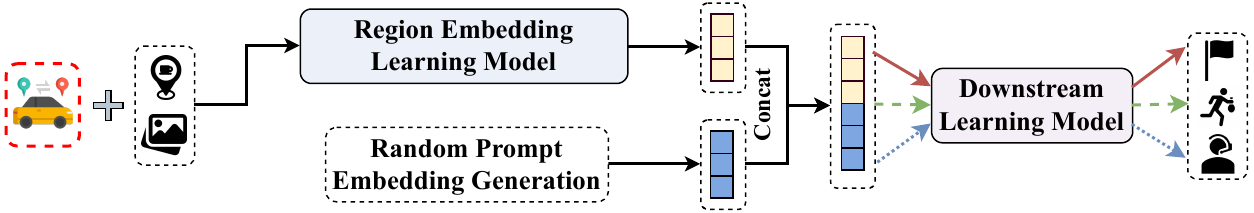}
        \vspace{-1mm}
        \caption{Representation learning with prompting}
        \label{fig:intro2}
    \end{subfigure}
    \begin{subfigure}[b]{\columnwidth}
        \vspace{2mm}
        \centering
        \includegraphics[width=0.95\columnwidth]{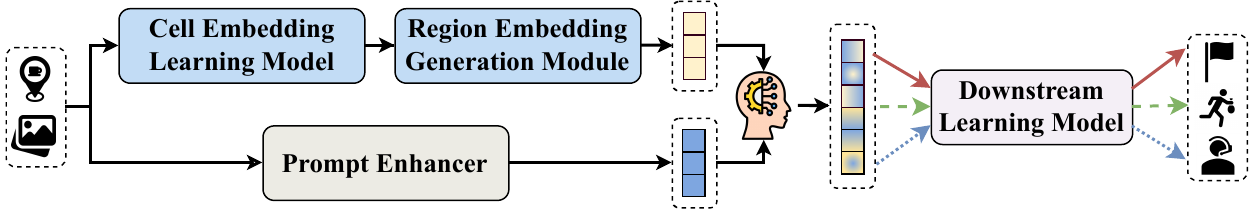}
        \vspace{-1mm}
        \caption{Ours: allowing flexible partitions and task-based prompting} 
        \label{fig:intro3}
    \end{subfigure}
    \vspace{-6mm}
    \caption{Region representation learning schemes.}
    \label{fig:introducation}
\end{figure}
\vspace{-1mm}

Recently, the use of multi-modal data for learning urban region representations has gained attention. A critical aspect of this process is the selection of input data features, often referred to as region features, where each type of features depicts a region from a distinct \emph{view}.
Existing studies commonly utilize human mobility data~\cite{HAFusion, MGFN, HDGE, ZE-Mob, CDAE, MP-VN, CGAL, ReMVC, DLCL, MVURE, ReCP, RAW, Region2Vec, CGAP, MuseCL} and POIs~\cite{HAFusion, MP-VN, CGAL, ReMVC, MVURE, ReCP, RAW, Region2Vec, CGAP}. 
Among these, the ones using human mobility data often demonstrate superior performance, as such data offer critical insights into movement pattern and hence functional relationships between regions.

Moreover, existing studies typically follow a  two-stage process, as shown in Fig.~\ref{fig:intro1}, with a generic region embedding learning stage and a downstream task learning stage. The first stage learns the embeddings for a set of predefined regions with all input region feature data, while the second stage trains a (separate) machine learning model for downstream tasks, e.g., crime count prediction, using the embeddings as the model input.

However, there are  three limitations in the existing studies:

\textbf{Limitation 1. Existing methods heavily depend on mobility data and underutilize publicly accessible data.}
Human mobility data plays a critical role in learning effective region embeddings~\cite{HAFusion, MGFN, ReCP, MVURE, ReMVC}.
However, their limited availability, particularly in underdeveloped regions, together with privacy issues, prevent models using such data from a wider adoption. 
Recently, several studies~\cite{RegionDCL, PG-SimCLR, urbanclip, MuseCL} leverage features from publicly accessible data (e.g., POIs from OpenStreetMap) to enhance model applicability. However, these studies suffer from the effectiveness of the learning models. 
Their learned embeddings have reported lower accuracy for downstream tasks comparing with those learned by the mobility-based models.

\textbf{Limitation 2. Existing studies lack the flexibility to utilize different urban features for different downstream tasks.}
Existing studies simply use all input features to learn region embeddings together, without considering their relevance to  specific downstream tasks.
During the downstream task learning stage, most studies directly use the region embeddings for downstream tasks without any adaptation for task-specific needs, as shown in Fig.~\ref{fig:intro1}. 
Prompt learning presents opportunities to incorporate task-specific adaptations into the region embeddings.
HREP~\cite{HREP} first attempted this idea (see  Fig.~\ref{fig:intro2}). It simply applies random prompt embeddings for downstream tasks, which fails to capture the correlation between features and downstream tasks.

\textbf{Limitation 3. Existing studies lack the flexibility to adapt to different formation of regions}
Existing studies typically rely on a single, predefined region formation for all downstream tasks, making it difficult to accommodate different downstream tasks with different region formations (or analytical tasks to explore different region formations). 
For example, population estimation may need to be done at the census tract level, whereas transportation planning concerns more on traffic-related region partitions. As a third example, real estate investors or house seekers may be more interested in regions defined by school zones. Existing studies will need to compute a different set of embeddings for each of these application scenarios, which is costly and less flexible. 

We summarize existing works for the issues above in Table~\ref{tab:related_work}. 
\vspace{-1mm}
\begin{table}[ht] \scriptsize
\centering
\caption{\small Comparison between Region Embedding Learning Methods}
\label{tab:related_work}
\renewcommand{\arraystretch}{1.2}
\setlength{\tabcolsep}{2pt}
\resizebox{0.85\columnwidth}{!}{
\begin{tabular}{cccc}
\toprule
\textbf{Models} & \multicolumn{1}{c}{\textbf{\begin{tabular}[c]{@{}c@{}}Publicly \\ accessible data\end{tabular}}} &
\multicolumn{1}{c}{\textbf{\begin{tabular}[c]{@{}c@{}}Prompting\end{tabular}}} &
\multicolumn{1}{c}{\textbf{\begin{tabular}[c]{@{}c@{}}Adaptive region \\ embeddings\end{tabular}}} 
\\ \midrule 
  \makecell[c]{\cite{MGRL4RE, DLCL, MP-VN, ReCP, HAFusion, HDGE} \\  \cite{CDAE,MGFN, CGAP, ZE-Mob, MuseCL, ReMVC,MVURE, RAW, CGAL}}& & & \\
\hline
 \cite{HREP}  &  & \checkmark & \\
\hline
\cite{urban2vec, m3g, urbanclip, RegionDCL, PG-SimCLR, MMGR, HGI} & \checkmark & &  \\
\hline
\cite{GeoVectors, cityFM} & \checkmark & & \checkmark  \\
\hline
\textbf{\model\ (ours)} & \ding{52} & \ding{52} & \ding{52}  \\
  
\bottomrule
\end{tabular}
}
\end{table}
\vspace{-1mm}

To address the issues above, we propose \model\ (Fig.~\ref{fig:intro3}),  a \underline{Flexi}ble model for urban \underline{Reg}ion representation learning.  
It takes a three-stage learning process that enables a flexible use of urban features to generate region embeddings tailored for different region formations and downstream tasks. 
\model\ is flexible in all three aspects discussed above:

(1)~It leverages urban region features from publicly accessible data, including POIs, land use data, satellite imagery, and street view imagery, which have wider availability than human mobility data. To effectively exploit  these features, we partition an area of interest into finer-grained spatial partition units using a hexagonal grid. We propose a novel \emph{multimodal \ul{grid} cell embedding \ul{learn}ing} (\gridGen) module and an environment context-based contrastive learning technique to capture distinctive urban patterns from each type of input feature and spatial correlations between different types of features, respectively. (Addressing Limitation~1)

(2)~It takes a three-stage learning process. The first two stages learn fine-grained grid cell embeddings and  
aggregate them into region embeddings, respectively. We propose an \emph{\ul{adap}tive \ul{region} embedding \ul{gen}eration} (\regionGen) module for the aggregation stage, which weighs the embeddings for the cells by their  overlapping areas with a region. Notably, this aggregation process is flexible, allowing grid cell embeddings to be combined into region embeddings regardless of the region partitioning methods. (Addressing Limitation~3)

(2)~It has a prompt learning process for its third stage, which enables it to flexibly utilize different types of features for different downstream tasks. We propose a novel \emph{\ul{prompt} \ul{enhancer}} (\prompt) module to tailor region embeddings for downstream tasks by integrating textual and street-view imagery features. 
To capture task-relevant information, \prompt\ consists of a \emph{\ul{t}ext\ul{-r}egion \ul{align}ment} (T-RAlign) module 
and a \emph{\ul{s}treet \ul{v}iew\ul{-r}egion \ul{align}ment} (SV-RAlign) module. 
T-RAlign incorporates task-specific geographic knowledge into region embeddings using dimension-wise similarity, while SV-RAlign extracts task-relevant visual features through adapted attention mechanisms. (Addressing Limitation~2)

To summarize, this paper makes the following contributions:

(1)~We propose a model named \model\ to generate effective and flexible region representations that can be adapted for different downstream tasks by leveraging publicly accessible data.

(2)~We propose a multimodal grid cell embedding learning module, followed by an adaptive region embedding generation module to generate region embeddings when a set of regions is given. These two modules capture urban patterns within grid cells and model their correlations to enhance region representation learning.

(3)~We propose a prompt enhancer module to tailor region embeddings for downstream tasks by effectively extracting task-relevant information from additional features and seamlessly integrating them into the embeddings.

(4)~We conduct extensive experiments to evaluate \model\ on five real-world datasets. The results show that \model\ outperforms all competitors, including those utilizing publicly accessible data and those based on human mobility data, across four downstream tasks (crime, check-in, service call, and population count predictions), by up to 202\% in term of accuracy.

\vspace{-2mm}
\section{Solution Overview}
\label{sec:solutionOverview}

This section presents the problem and model overview. 

\begin{figure*}[htbp]
  \centering
  \includegraphics[width=\textwidth]{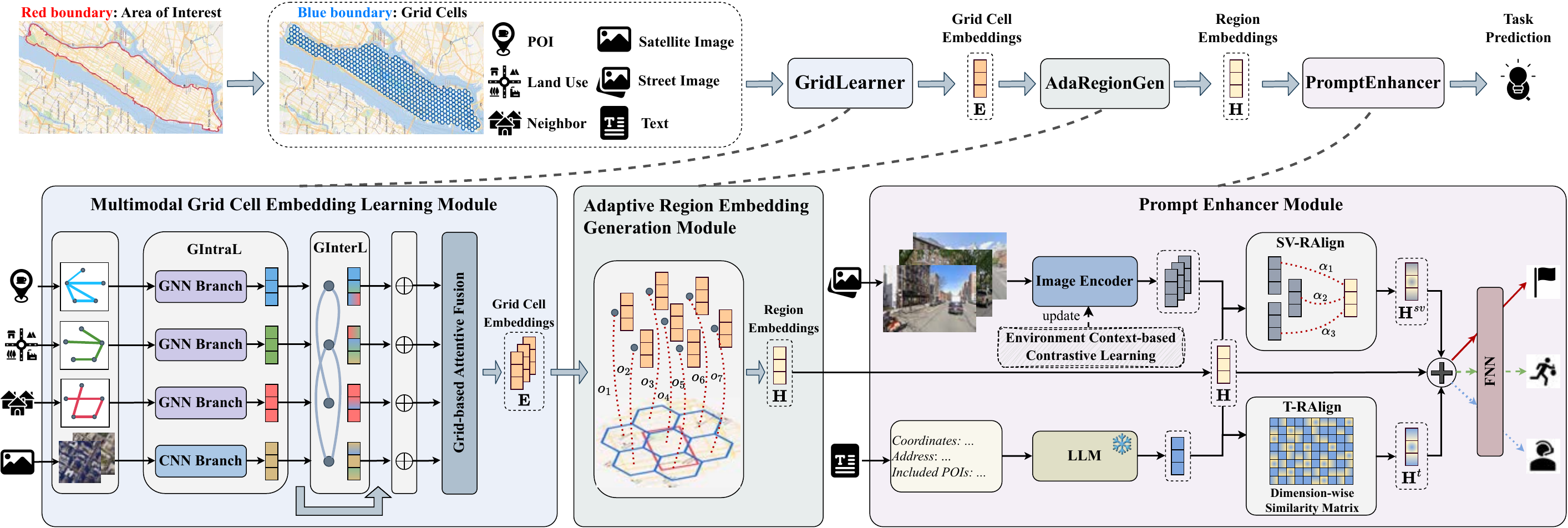}
  \vspace{-6mm}
  \caption{\model\ model overview. The model processes a set $C$ of grid cells, each with associated features, through a three-stages learning process to generate flexible region embeddings to accommodate the needs of different downstream tasks with different region formations.: (1)~\gridGen\ computes fine-grained grid cell embeddings $\mathbf{E}$. (2) \regionGen\ aggregates fine-grained grid cell embeddings to produce region embeddings $\mathbf{H}$, given an input partition of regions. (3) \prompt\ further tailors the region embeddings with additional features as guided by given downstream tasks.}
  \label{fig:model_overview}
\end{figure*}

\textbf{Problem statement.} 
Given a spatial area of interest with publicly accessible features (detailed in Section~\ref{subsubsec:feature_collection}) and a set of non-overlapping regions $R$ in this area, we aim to learn an embedding function $f: r_i \rightarrow \mathbf{h}_i$ that maps a region $r_i \in \mathcal{R}$  to a $d$-dimensional vector $\mathbf{h}_{i}$. The learned embeddings are expected to be applicable in different downstream tasks. Then, for each downstream task (e.g., crime count prediction), we learn a prediction function $g: \mathbf{h}_i \rightarrow y_{i}$, where $y_{i} \in \mathbb{R}$ is typically a numerical indicator for the task.


\textbf{Model overview.}
Fig.~\ref{fig:model_overview} shows the overall structure of our model  \model, which consists of three main learning stages.
(1)~\model\ takes a set of grid cells as the input. The grid cells come from a fine-grained partitioning over the spatial area of interest that we perform as part of data preparation (Section~\ref{subsec:data_preparation}). \model\ learns the embeddings of grid cells across different features through a Multimodal \ul{Grid} Cell Embedding \ul{Learn}ing module (\gridGen, Section~\ref{subsec:gridEmbLeaning}). 
(2) Then, the \ul{Adap}tive \ul{Region} Embedding \ul{Gen}eration module (\regionGen) aggregates the fine-grained cell embeddings to generate region embeddings for  the input regions (Section~\ref{subsec:regionEmbGen}). Starting from the fine-grained cell embeddings allows \model\ to flexibly adapt to different sets of regions which may come from different downstream tasks (or analytical tasks to explore different way to form regions). 
(3) Finally, the Prompt Enhancer module (\prompt) refines the generic region
embeddings with extra features, guided by given tasks (Section~\ref{subsec:promptEnhancer}).

\section{Proposed Model}
\label{sec:model}

This section details the \model\ model. We summarize the frequently used symbols in Table~\ref{tab:symbols}.

\begin{table}[ht]
\centering
\caption{Frequently Used Symbols}
\label{tab:symbols}
\renewcommand{\arraystretch}{1.1}
\resizebox{\columnwidth }{!}{
    \begin{tabular}{cl}
    \toprule
    \textbf{Symbol} & \textbf{Description} \\ 
    \hline
     $S$ & A spatial area of interest  \\ \hline
    $R$ & A set of regions (non-overlapping space partitions)  \\ \hline 
    $n$ & The number of regions \\ \hline
    $C$ & A set of spatial grid cells (basic space partition units) \\ \hline 
    $m$ & The number of grid cells \\ \hline
     $\mathbf{p}_i$ & POI feature of cell $c_i$ \\ \hline 
    $\mathbf{l}_i$ & Land use feature of cell $c_i$ \\ \hline 
    $\mathbf{gn}_i$ & Geographic neighbor feature of cell  $c_i$ \\ \hline 
    $\mathbf{si}_i$ & Satellite imagery feature of cell  $c_i$ \\ \hline 
    $\mathbf{sv}_i$ & Street view imagery feature of cell  $c_i$ \\ \hline 
    $\mathbf{t}_i$ & Textual feature of cell  $c_i$  \\ \hline 
    $\mathbf{E}$ & The embeddings of grid cells \\ \hline 
    $\mathbf{H}$ & Adaptive region embeddings \\  
    \bottomrule
    \end{tabular}
}
\vspace{-4mm}
\end{table}

\subsection{Data Preparation}
\label{subsec:data_preparation}

\subsubsection{Grid Cell Construction}
\label{subsubsec:grid_construction}
We partition the input spatial area of interest into a set $C$ of grid cells, where $c_i$ denotes the $i$-th cell. Here, the grid cells are supposed to be finer-grained spatial partitions than the regions, allowing for flexible formations of regions as required by downstream tasks later on. We partition the area using a hexagonal grid, as illustrated in Fig.~\ref{fig:model_overview} (the blue grid on the map shown at the top left), which provides several advantages.
First, cells (which are of a small size) mitigate spatial heterogeneity by enabling localized feature learning, allowing \model\ to effectively capture local variations within a region.
Second, hexagonal cells in particular offer more uniform coverage than cells of other shapes (e.g., squares), as each cell is surrounded by six equidistant neighbors.
Third, hexagonal cells are easier to approximate natural boundaries, improving spatial coverage and making them ideal for regions with irregular boundaries~\cite{hexGrid1, hexGrid2}.

\subsubsection{Feature Preparation}
\label{subsubsec:feature_collection}

\begin{figure}[htbp]
  \centering
  \includegraphics[width=\columnwidth]{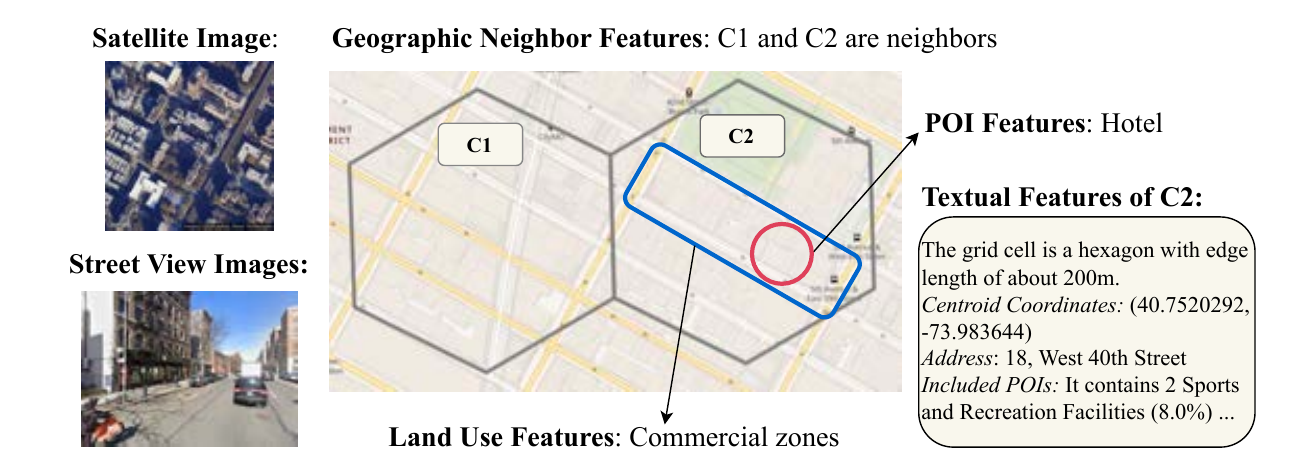}
  \caption{Example of input cell features.}
  \label{fig:features_example}
\end{figure}

We use six types of features for each cell, which are all publicly accessible, as shown in the Fig.~\ref{fig:features_example}

\textbf{POI features.}
For each cell, we count the number of POIs that belong to one of 15 POI categories \emph{(educational institutions, commercial and industrial properties, accommodation, cultural and recreational venues, healthcare and medical facilities, entertainment venues, places of worship, food and drink establishments, parking facilities, transportation and transit facilities, residential properties, camping and outdoor recreation sites, sports and recreation facilities, financial services, and others.)} from OpenStreetMap~\cite{osm} as the POI feature. We denote the POI feature of  cell $c_i$ as 
$\mathbf{p}_i \in \mathbb{R}^{15}$. 

\textbf{Land use features.} 
Similar to POI features, we count the numbers of zones that belong to 20 different land use types \emph{(grass, park, cemetery, forest, scrub, meadow, farmland, industrial, heath, retail, military, nature reserve, residential, commercial, orchard, farmyard, allotments, recreation ground, vineyard, and quarry.)} within a cell. The land use feature of cell $c_i$ is denoted as $\mathbf{l}_{i} \in \mathbb{R}^{20}$. 

\textbf{Geographic neighbor features.}
This feature indicates the adjacency relationships between cells. We use $\mathbf{gn}_{i} \in \mathbb{R}^{6}$ to denote a vector of the six direct neighboring cells of $c_i$.

\textbf{Satellite imagery features.}
Satellite images capture rich coarse-grained urban patterns of grid cells. We use $\mathbf{si_i} \in \mathbb{R}^{H \times W \times 3}$ to denote the satellite imagery feature of $c_i$, where $H$ and $W$ denote the height and the width of the satellite image of $c_i$, respectively. 

\textbf{Street view imagery features.}
Street view images capture finer-grained urban patterns.
We use $\mathbf{sv}_i = \{\mathbf{sv}_{i,1}, \mathbf{sv}_{i,2}, \cdots \}$ to denote the set of street view images captured within the area of $c_i$, where each image is also in the shape of $\mathbb{R}^{H \times W \times 3}$.
Note that different cells may have different numbers of street view images. To collect the street view images, we begin by sampling geocoordinate points along the road network at 100-meter intervals. Next, we remove redundant points that are within 20-meter to one another. In cases where the number of sampled points is less than 5 within a grid cell, we randomly sample additional points for that cell. Finally, we collect street view images from four different directions (0\textdegree, 90\textdegree, 180\textdegree, 270\textdegree) at each sampled point.

\textbf{Textual features.}
We generate textual features for each cell by describing them from different aspects in text, including geometric properties, addresses, and POIs within them, to enable learning the urban features from a semantic perspective. 
We denote the textual feature of $c_i$ as $\mathbf{t}_{i} \in \mathbb{R}^{S}$, where $S$ refers to the maximum length of a textual description. We elaborate this feature in Section~\ref{subsubsec:textregionAlign}.

We use the first four features to learn cell embeddings, capturing their functionality, spatial structure, correlations, and urban patterns, while the last two features will be used later to tailor region embeddings for downstream tasks. 
POIs and land use categories reflect the functional roles of urban areas; geographic neighbors reflect spatial relationships; satellite images provide visual insights into the physical layout and urban patterns. These features are generic for urban representation learning. 
In contrast, textual data captures nuanced details such as the presence  of a large number of ``entertainment venues'',  which may correlate with and suit a common downstream task, crime prediction. Street view images offer ground-level context, such as building density, which is crucial for tasks such as population prediction. These features are more suitable for task-relevant adaptation of the region embeddings. 

\subsection{Multimodal Grid Cell Embedding Learning}
\label{subsec:gridEmbLeaning}
The \gridGen\ module learns cell embeddings through four \emph{views} each corresponding to a type of input features. It learns correlations between views and between cells, forming robust embeddings.

\subsubsection{Grid-based Intra-view Feature Learning}
\label{subsubsec:intra-view}
We leverage GNNs for POI, land use, and geographic neighbor features, and a CNN for satellite imagary features, to suit the different types of features.

\textbf{GNN branches.}
We construct feature-aware grid graphs on the POI ($p$), land use ($l$), and geographic neighbor ($gn$) features of cells, separately, to help capture the correlation between cells based on such features.
Let $\mathcal{G}^\mathcal{X} = (\mathcal{V}, \mathcal{E}, \mathbf{A}^\mathcal{X})$ be a grid  graph based on a specific feature $\mathcal{X}$, where $\mathcal{X} \in \{p, l, gn\}$.
Here, $\mathcal{V} = \{c_1, c2, \cdots, c_m\}$ denotes the set of $m$ vertices (i.e., $m$ grid cells in the input area of interest);  $\mathcal{E}$ denotes the set of edges between vertices; and 
$\mathbf{A}^\mathcal{X}$ is a weighted adjacency matrix, where $\mathbf{A}^\mathcal{X}_{i,j}$ is the cosine similarity between feature vectors of $c_i$ and $c_j$.

Once the feature-aware grid graphs are constructed, we employ Graph Attention Networks (GAT)~\cite{gat} to produce grid cell embeddings on each view of features. 
GAT stacks multiple graph attention layers to compute the correlation between vertices and aggregate vertex embeddings based on correlation scores. For a given graph attention layer at the $g$-th layer, its process is as follows. We omit 
the superscript $\mathcal{X}$ hereafter for simplicity when the context is clear.
{\small
\begin{align}
\alpha_{ij}^{g} & = \mathrm{Softmax}\left(\sigma\left(\mathbf{a}^\intercal\left(\mathbf{W}\mathbf{z}^{g}_{i} ||\mathbf{W}\mathbf{z}^{g}_{j} ||\mathbf{w}\mathbf{A}_{i,j}\right)\right)\right), \label{eq:gat_2} \\
\mathbf{z}^{g+1}_{i} & =  \sigma (\sum_{j \in [1, m]} \alpha_{ij}^{g} \mathbf{z}^{g}_{i}). \label{eq:gat_3}
\end{align}}
Here, $a_{ij}^{g}$ denotes the correlation (i.e., the  normalized correlation score) between $c_i$ and $c_j$ w.r.t. their embeddings $\mathbf{z}_i^g$ and $\mathbf{z}_j^g$ in the $g$-th GAT layer. We use $\mathbf{a} \in \mathbb{R}^{3d}$, $\mathbf{W} \in  \mathbb{R}^{d\times d}$, and $\mathbf{w} \in \mathbb{R}^{d}$ to denote learnable parameters, and $\sigma$ is the LeakyReLU activation function.  
The input to the $1$-st layer, $\mathbf{z}^{0}_{i}$, is obtained by random initialization. 

We apply three GATs to POI, land use, and geographic neighbor features separately to obtain representations for each feature view, denoted as $\mathbf{Z}^{p}$, $\mathbf{Z}^{l}$, and $\mathbf{Z}^{gn}$, each in the shape of $\mathbb{R}^{m\times d}$.

\textbf{CNN Branch.}
We employ ResNet~\cite{resnet} followed by an MLP to encode the satellite images of grid cells into embeddings:
{\small
\begin{equation}
\mathbf{z}^{si}_{i} = \mathrm{MLP}(\mathrm{ResNet}(\mathbf{si}_i)),
\end{equation}
}where $\mathbf{z}^{si}_{i}$ denotes the embedding of the satellite image of $c_i$, and the MLP is an additional projection layer. Further, $\mathbf{Z}^{si} \in \mathbb{R}^{m \times d}$ denotes the embeddings of cells on satellite imagery features.

\subsubsection{Grid-based Inter-view Feature Learning}
\label{subsubsec:inter-view} 

Next, we introduce a grid-based inter-view feature learning module to learn spatial correlations between grid cells across views.
To learn the cross interactions between different views of the same cell, we simply apply a one-layer self-attention network~\cite{attention} over 
$\mathbf{Z} \in \mathbb{R}^{4 \times m \times d}$ which stacks \emph{intra-view cell embeddings} $\mathbf{Z}^p$, $\mathbf{Z}^l$, $\mathbf{Z}^{gn}$,  and $\mathbf{Z}^{si}$. Now we denote $\mathbf{Z}$ and $\mathbf{Z}^{\mathcal{X}}$ as $\mathbf{Z}_{intra}$ and $\mathbf{Z}^{\mathcal{X}}_{intra}$, respectively, to distinguish with the embedding matrices used later for inter-view features, where $\mathcal{X} \in \{p, l, gn, si\}$.

Formally, given the intra-view cell embedding matrices $\mathbf{Z}_{intra}$, we compute the attention coefficients between different views via a self-attention module as follows: 
\begin{equation}\label{eq:attn_coeffients}
\mathbf{A}_{inter} = \mathrm{Softmax}\left(\frac{(\mathbf{W}_Q {\mathbf{Z}_{intra})} \cdot {(\mathbf{W}_K {\mathbf{Z}_{intra}})}^\intercal}{\sqrt{d}}\right),
\end{equation}
where $\mathbf{W}_Q \in \mathbb{R}^{d \times d}$ and $\mathbf{W}_K \in \mathbb{R}^{d \times d}$
are learned parameter matrices of the linear transformations, which transform $\mathbf{Z}_{intra}$ to form projected matrices in latent spaces; $\sqrt{d}$ is a scaling factor; and $\mathbf{A}_{inter} \in \mathbf{R}^{v \times v}$ is a coefficient matrix that records the correlation between every two views.
Multi-head attention~\cite{attention} is applied here to enhance the model learning capacity, while its details are omitted as it is a direct adoption.

After that, we compute hidden representations of the grid cells based on the attention coefficients $\mathbf{A}_{inter}$:
\begin{equation}\label{eq:attn_c}
    \mathbf{Z}_{inter} = \mathbf{A}_{inter} \cdot \left(\mathbf{W}_V {\mathbf{Z}_{intra}}\right).
\end{equation}
where $\mathbf{W}_V \in \mathbb{R}^{d' \times d'}$ is the same as $\mathbf{W}_Q$ and $\mathbf{W}_K$ above. We have obtained four embedding matrices corresponding to the four different input views, denoted as $\mathbf{Z}_{inter} = \{$$\mathbf{Z}_{inter}^p$, $\mathbf{Z}_{inter}^l$, $\mathbf{Z}_{inter}^{gn}$, $\mathbf{Z}_{inter}^{si}\}$. We call these matrices the \emph{inter-view cell embedding matrices}.

Finally, we adaptively combine the intra-view cell embedding matrices $\{\mathbf{Z}_{intra}^\mathcal{X}\}$ and the inter-view cell embedding matrices $\{\mathbf{Z}_{inter}^\mathcal{X}\}$ with a learnable weight $\beta \in [0,1]$, to form the view-based cell embeddings $\{\mathbf{Z}^\mathcal{X}\}$:
\begin{equation}
{\mathbf{Z}^\mathcal{X}} = \beta{\mathbf{Z}_{intra}^\mathcal{X}} + \left( 1-\beta \right){\mathbf{Z}_{inter}^\mathcal{X}}
\end{equation}


\subsubsection{Grid-based Dual-Feature Attentive Fusion} 
\label{subsubsec:fusionmodule}
We adopt the dual-feature attentive fusion module (DAFusion) from HAFusion~\cite{HAFusion} to generate the final cell embeddings $\mathbf{E}$ based on the cell embeddings $\mathbf{Z}$. DAFusion consists of two sub-modules: view-aware attentive fusion (ViewFusion) and region-aware attentive fusion (We call it cell-aware attentive fusion, denoted as CellFusion, since we are using this module at the cell level).

\textbf{View-aware Attentive Fusion.} 
ViewFusion leverages the attention mechanism to learn  \emph{fusion weights} of the views to aggregate the view-based cell embeddings $\{\mathbf{Z}^p, \mathbf{Z}^l, \mathbf{Z}^{gn}, \mathbf{Z}^{si}\}$. It first computes correlation scores between different views as follows:
\begin{equation}
a^{kj}_{i} = \mathrm{LeakyReLU}\left(\mathbf{a}^\intercal\left(\mathbf{W}\mathbf{z}^k_{i}||\mathbf{W}\mathbf{z}^j_{i}\right)\right),
\end{equation}
where $a^{kj}_{i}$ is the correlation score between the $j$-th and the $k$-th views of grid cell $c_i$.

Then, we aggregate the correlation scores along the views and the cells to obtain an overall weight for each view. Afterwards, we apply a Softmax function to obtain the normalized fusion weight $\alpha^k$ ($k \in [1, 4]$) of each view.
\begin{equation}
\alpha^k = \mathrm{Softmax}\left(\frac{1}{m} \sum^{m}_{i = 1} \sum^{4}_{j=1} a_{i}^{kj} \right),
\end{equation}

We use the fusion weights to fuse the view-based cell embeddings into a single embedding matrix, denoted as $\widetilde{\mathbf{Z}}$:

\begin{equation}
\widetilde{\mathbf{Z}} = \sum_{k=1}^{4}\alpha^k \cdot \mathbf{Z}^k
\end{equation}

\textbf{Cell-aware Attentive Fusion.} 
CellFusion further applies self-attention on the embeddings $\widetilde{\mathbf{Z}}$ learned by
ViewFusion, to encode the higher order correlations among the learned cell embeddings.
Embeddings $\widetilde{\mathbf{Z}}$ are first fed into a self-attention module to produce the hidden representations $\hat{\mathbf{Z}}$ of the grid cells (which resemble Equations~\ref{eq:attn_coeffients} and \ref{eq:attn_c}).

Then, $\hat{\mathbf{Z}}$ is combined with $\widetilde{\mathbf{Z}}$ via a residual connection, followed by layer normalization (LayerNorm) and dropout. Afterward, an MLP is applied, along with another layer normalization and residual connection, to further enhance the model’s learning capacity, as expressed below:
\begin{gather}
    \hat{\mathbf{Z}}' = \mathrm{LayerNorm}\left(\widetilde{\mathbf{Z}} + \mathrm{Dropout} \left(\hat{\mathbf{Z}}\right)\right), \label{eq:attn_postprocess1} \\
    {\mathbf{E}} = \mathrm{LayerNorm}\left(\hat{\mathbf{Z}}' + \mathrm{Dropout}\left(\mathrm{MLP}\left
    (\hat{\mathbf{Z}}'\right)\right)\right). \label{eq:attn_postprocess2}
\end{gather}
Here, $\hat{\mathbf{Z}}'$ is the output of the first layer normalization and $\mathbf{E}$ is the output cell embeddings.

We stack multiple layers of the aforementioned structure, with the output from the final layer serving as our learned cell embeddings, denoted as $\mathbf{E}$.


\subsubsection{Module Training.}
\label{subsubsec:modeltrain}
We leverage a multi-task learning objective $\mathcal{L}$ to learn the cell representations, which consists of four sub-objective functions, each corresponding to a type of features.

Given embeddings $\mathbf{E}$, we first generate feature-oriented cell embeddings $\mathbf{E}^\mathcal{X}$ for feature $\mathcal{X}$ (now  $\mathcal{X}$ denotes one of the four types of input features above) by adopting an MLP, which can be represented as $\mathbf{E}^\mathcal{X} = \mathrm{MLP}_{\mathcal{X}}(\mathbf{E})$. As a result, we obtain four types of feature-oriented embeddings $\mathbf{E}^\mathcal{X}$, each using a different objective.

\textbf{POI-Oriented Objective}
We utilize the graph reconstruction task to reconstruct the POI adjacent matrix $\mathbf{A}^p$ using the POI task embeddings $\mathbf{E}^p = \{e^p_i\}^{m}_{i=1}$. The objective function $\mathcal{L}^p$ is formulated as follows:
\begin{equation}\label{eq:sub_objective_poi}
    \mathcal{L}^p = \frac{1}{m}\frac{1}{m}\sum_{i=1}^{m}\sum_{j=1}^{m}\Big|\mathbf{A}^p_{i,j} - {\mathbf{e}^{p}_{i}} \cdot \mathbf{e}^p_{j}\Big|,
\end{equation}
where vectors $\mathbf{e}^{p}_{i}$ and  $\mathbf{e}^p_{j}$ are the learned embeddings of grid cells $c_i$ and $c_j$ mapped towards POI features, and their dot product represents the cell similarity in the embedding space. The intuition is that the learned embeddings should reflect the cell similarity as entailed by the input features. 

\textbf{Land Use-Oriented Objective}
The land use objective mirrors the POI objective, aiming to reconstruct the land use adjacent matrix $\mathbf{A}^l$ using the land use-oriented embeddings $\mathbf{E}^l$. The objective function $\mathcal{L}^p$ is defined as follows:
\begin{equation}\label{eq:sub_objective_landuse}
    \mathcal{L}^l = \frac{1}{m}\frac{1}{m}\sum_{i=1}^{m}\sum_{j=1}^{m}\Big|\mathbf{A}^l_{i,j} - {\mathbf{e}^{l}_{i}} \cdot \mathbf{e}^l_{j}\Big|,
\end{equation}
where vectors $\mathbf{e}^{l}_{i}$ and  $\mathbf{e}^l_{j}$ are the learned embeddings of grid cells $c_i$ and $c_j$ mapped towards land use features.

\textbf{Geographic Neighbor-Oriented Objective}
By the First Law of Geography~\cite{firstlawofGeo}, nearby grid cells are likely to have similar functionality and embeddings. Accordingly, we utilize the triplet loss~\cite{tripletloss} as the geographic neighbor objective function $\mathcal{L}^{gn}$ based on $\mathbf{E}^{gn}$. This loss aims to minimize the distance between an anchor and its positive sample while maximizing the distance between the anchor and a negative sample, ensuring a margin of separation. Formally, given the task embeddings $\mathbf{E}^{gn}$, the geographic neighbor objective function is formulated as:
\begin{equation}\label{eq:sub_objective_neighbor}
    \mathcal{L}^{gn} = \frac{1}{m}\sum_{i=1}^{m} \mathrm{max} 
    (\Vert \mathbf{e}_i - \mathbf{e}_i^{(pos)} \Vert_{2} - 
    \Vert \mathbf{e}_i - \mathbf{e}_i^{(neg)} \Vert_{2} + \mathrm{M}
    , 0),
\end{equation}
where $\mathbf{e}_i^{(pos)}$ represents a positive sample, which corresponds to a geographic neighbor of grid cell $c_i$. Conversely, $\mathbf{e}_i^{(neg)}$ denotes a negative sample, representing a non-geographic neighbor. We use $\mathrm{M}$ to denote a predefined margin that ensures sufficient separation between positive and negative pairs. The term $\Vert \cdot \Vert_2$ denotes the L2 (i.e., Euclidean) distance between embeddings.

\textbf{Satellite Image-Oriented Objective}
We leverage the object counting task to predict the total number of POIs within the satellite image of each grid cell. Satellite images contain POI information, such as buildings, roads, farmlands, and shops. This task encourages the model to extract POI features from the satellite images.

We obtain the ground-truth POI count $y_i$ for grid cell $c_i$ by summing up its POI feature vector $\mathbf{p}_i$, i.e., $y_i = \mathrm{sum}(\mathbf{p}_i)$.
Subsequently, the predicted POI count $\hat{y_i}$ is computed by passing the corresponding task embedding $\mathbf{e}^{si}_i \in \mathbf{E}^{si}$ through an MLP, expressed as $\hat{y_i} = \mathrm{MLP}(\mathbf{e}^{si}_i)$.
Given $y_i$ and $\hat{y_i}$, we employ the smooth L1 loss~\cite{smoothl1loss} as the satellite image objective function $\mathcal{L}^{si}$, computed as follows:
\begin{gather}
    \mathcal{L}^{si} = \frac{1}{m}\sum_{i=1}^{m} u_i \label{eq:sub_objective_SI_1} \\
    u_i = 
    \left\{
    \begin{array}{lr}
         0.5 \cdot (\hat{y_i} - y_i)^2 & if |\hat{y_i} - y_i| < \beta,  \\
         \beta \cdot |\hat{y_i} - y_i| - 0.5 \cdot \beta^2 & otherwise.
    \end{array}
    \right. 
    \label{eq:sub_objective_SI_2}
\end{gather}
where $\beta$ is a threshold hyperparameter ($\beta$ = 1 in our experiments). When the absolute difference between the prediction and the ground truth is smaller than $\beta$, the loss behaves like the L2 loss (quadratic), ensuring smooth gradients for small errors. Conversely, if the difference exceeds $\beta$, the loss becomes an L1 loss (linear), which reduces the influence of outliers and enhances robustness.




Finally, the overall objective function is derived by summing up the feature-oriented objective functions as follows:
{\small
\begin{equation}
    \mathcal{L} = \mathcal{L}^{p} + \mathcal{L}^{\mathcal{L}} + \mathcal{L}^{gn} + \mathcal{L}^{si} 
\end{equation}
}

\subsection{Adaptive Region Embedding Generation}
\label{subsec:regionEmbGen}
Next, we generate the region embeddings $\mathbf{H} = \{h_{i}\}^{n}_{i=1}$ by aggregating the embeddings $\mathbf{E}$ of grid cells corresponding to input regions based on their spatial locations. 
Given a region $r_j$, we first find a set of grid cells, denoted as $\mathcal{C}_{r_j} = \{ c_1, \cdots, c_i, \cdots \}$, where each cell $c_i$ either spatially intersects with or is contained within $r_j$.
In addition, we compute the overlapping coefficient between  $r_j$ and each $c_i \in \mathcal{C}_{r_j}$ based on their areas, which indicates the relative importance of $c_i$ to $r_j$, as follows:
{\small
\begin{equation}\label{eq:aggregation_1}
    o_{{r_j} \cap {c_i}} = \frac{\mathrm{Area}(r_j \cap c_i)}{\mathrm{Area}(c_i)},
\end{equation}
}where $\cap$ denotes the spatial intersection, and $\mathrm{Area}(\cdot)$ computes the size of a given spatial area.
Then, we fuse the cell embeddings with their overlapping coefficients to region $r_j$ and generate the region embedding $\mathbf{h}_j$:
{\small
\begin{equation}\label{eq:aggregation_2}
\mathbf{h}_j = \sum_{c_i \in \mathcal{C}_{r_j}} o_{{r_j} \cap {c_i}} \cdot \mathbf{e}_i,
\end{equation}
}Here, $\mathbf{e}_i \in \mathbf{E}$ is the cell embedding of $c_i$.

\vspace{-2mm}
\subsection{Prompt Enhancer for Task Learning}
\label{subsec:promptEnhancer}

We propose a \emph{prompt enhancer} (\prompt)  based on prompt learning, which refines the general region embeddings learned above for better adaptability across downstream tasks. \prompt\ integrates complementary features--textual descriptions and street view images--to provide rich contextual information, aligning region embeddings with task-specific demands for more accurate predictions.  It consists of two modules: the \emph{text-region encoding} module, which encodes semantic insights from textual descriptions, and the \emph{street view-region encoding} module, which incorporates ground-level visual details from street view images.

\vspace{-2mm}
\subsubsection{Text-Region Encoding}
\label{subsubsec:textregionAlign}
The text-region encoding module consists of three main steps: cell description generation, region embedding generation, and text-region embedding alignment.

\textbf{(1) Cell description generation.}
We developed a textual description template for cells based on the POI information, which serves as the prompt to effectively extract geographic knowledge from LLMs. The template includes the following key information of a cell:
(1)~\textit{geometric properties} describing the shape and size of a grid cell. 
(2)~\textit{address} referring to the detailed street address of the POI located at the center of a cell
(3)~\textit{POI information} including the categories and numbers of POIs within a cell.

\begin{figure}[htbp]
  \centering
  \includegraphics[width=0.9\columnwidth]{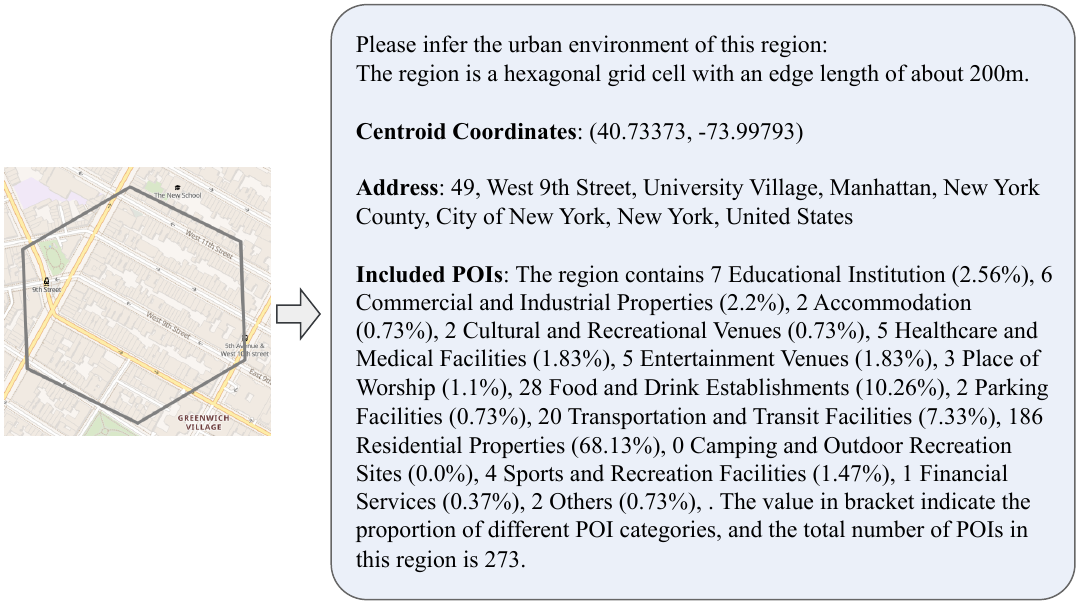}
  \caption{Example of the textual description of a grid cell.}
  \label{fig:textual_discription_example}
\end{figure}

An example of the generated textual description of a cell is shown in Fig.~\ref{fig:textual_discription_example}. We collect POI information from OpenStreetMap and generate detailed street address using the reverse geocoding functionality of the Nominatim API~\cite{Nominatim}.

\textbf{(2) Region embedding generation.}
After obtaining the cell descriptions, we generate their embeddings using a pre-trained parameter-frozen LLM (we use Llama 3 8B Instruct in our experiments)~\cite{Llama}.  
The input textual descriptions of all cells $\mathbf{T} \in \mathbb{R}^{m \times S}$ ($S$ refers to maximum length of a  textual cell description) are first tokenized and then processed into embeddings. We use the last token embeddings from the last hidden layer of the LLM as the final text embeddings of grid cells, denoted as $\mathbf{E}^{t} \in \mathbb{R}^{m \times d_{llm}}$, since the last tokens  capture information from all preceding tokens~\cite{linnot}. 
Here, $d_{llm}=4096$ is the dimensionality of the text embeddings. Note that using the frozen LLM parameters has the benefit of preserving the intrinsic geographic knowledge  learned by the LLM.

We use the same cell-to-region embedding aggregation approach as described in Section~\ref{subsec:regionEmbGen} to obtain the region embeddings from textual features, i.e., following Equations~\ref{eq:aggregation_1} and ~\ref{eq:aggregation_2}. The textual embeddings of regions are denoted as $\mathbf{H}^{t} \in \mathbb{R}^{n \times d_{llm}}$.

\textbf{ (3) Text-region embedding alignment.}
To integrate the semantic insights from textual embeddings with the region embeddings learned earlier, we design a \emph{text-region alignment} (T-RAlign)  module using dimension-wise similarity computation. This module extracts task-relevant geographic knowledge from the textual embeddings, with guidance given by a downstream task.

\begin{figure}[htbp]
  \centering
  \includegraphics[width=\columnwidth]{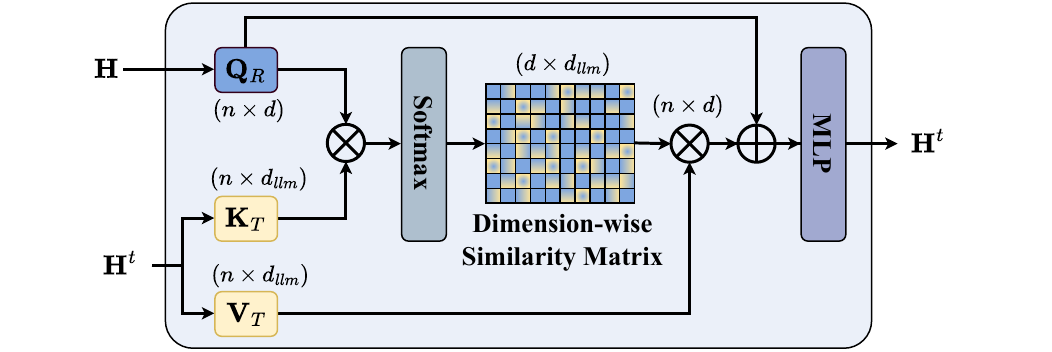}
  \caption{Text-region alignment module}
  \label{fig:text_region_align}
\end{figure}

As shown in Fig.~\ref{fig:text_region_align}, we employ linear transformations on $\mathbf{H}^{t}$ and $\mathbf{H}$ to form three projected matrices in latent space: $\mathbf{Q}_R \in \mathbb{R}^{n \times d} = \mathbf{W}_Q\mathbf{H}$, $\mathbf{K}_T \in \mathbb{R}^{n \times d_{llm}} = \mathbf{W}_K\mathbf{H}^{t}$, $\mathbf{V}_T \in \mathbb{R}^{n \times d_{llm}} = \mathbf{W}_V\mathbf{H}^{t}$. Here, $\mathbf{W}_Q$, $\mathbf{W}_K$, and $\mathbf{W}_V$ are learned parameters. 
Next, we compute the dimension-wise similarity matrix as follows:
{\small
\begin{equation}
    \mathbf{M}^{t} = \mathrm{Softmax} \left(\left(\mathbf{W}_Q\mathbf{H}\right)^{T} \left(\mathbf{W}_K\mathbf{H}^{t}\right)\right),
\end{equation}
}where $\mathbf{M}^{t} \in \mathbb{R}^{d \times d_{llm}}$ captures the similarity between the dimensions of the two embeddings. 

Then, we compute the retrieved textual embeddings by applying dimension-wise feature aggregation via matrix multiplication between $\mathbf{M}^{t}$ and $\mathbf{V}_T$. The embeddings are then combined with the input region embeddings $\mathbf{H}$ using element-wise addition. Finally, the result is passed through an MLP to update the output embeddings $\mathbf{H}^{t} \in \mathbb{R}^{n \times d}$. Formally, this process is expressed as:
{\small
\begin{equation}
    \mathbf{H}^{t} = \mathrm{MLP} \left(\left(\left(\mathbf{W}_V\mathbf{H}^{t}{\mathbf{M}^{t}}^{T}\right) + \mathbf{H}\right)\right).
\end{equation}
}

Through this text-region embedding alignment, we transfer the geographic knowledge encoded with the LLM into the region embeddings, thereby enhancing \model’s overall performance.

\subsubsection{Street View-Region Encoding}
\label{subsubsec:SVregionAlign}
The street view-region encoding module contains two main steps: street view image embedding learning, and street view-region embedding alignment. 

\begin{figure}[htbp]
  \centering
  \includegraphics[width=0.9\columnwidth]{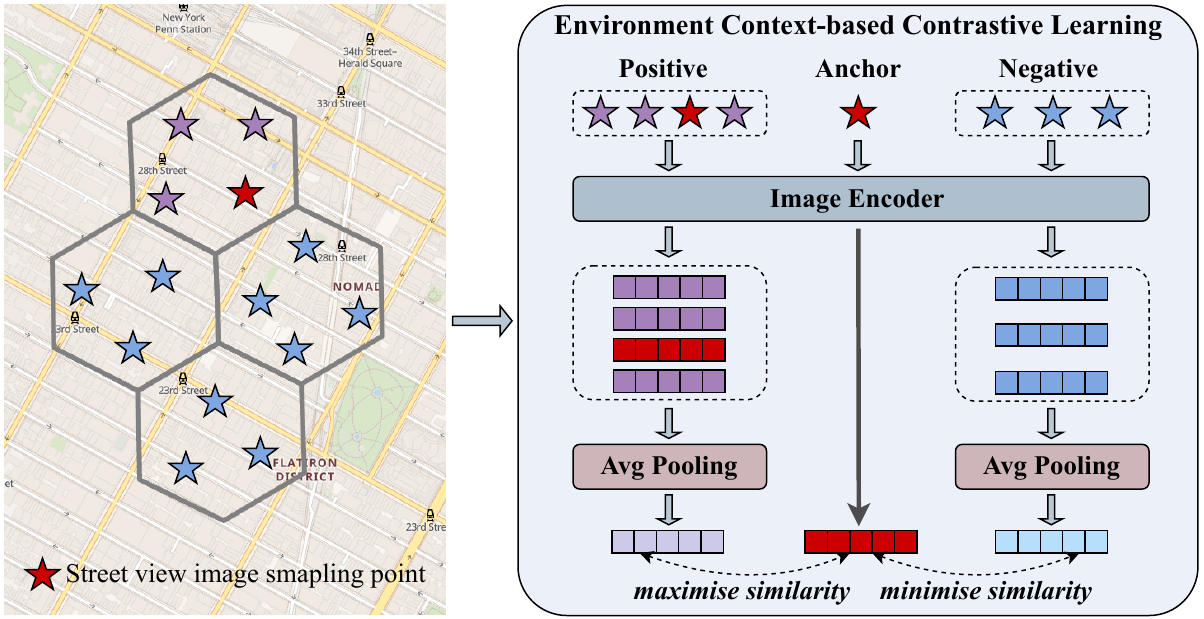}
  \caption{Street view image embedding learning}
  \label{fig:sv_cl}
\end{figure}

\textbf{(1) Street view image embedding learning.}
We propose an environment context-based contrastive learning approach to learn the representation of street view images as illustrated in Fig.~\ref{fig:sv_cl}. Motivated by the observation that street view images from the same cell exhibit strong correlations, we aim to maximize the similarity between a street view image and its corresponding  cell's environmental context while minimizing its similarity with unrelated cells. This ensures the learned embeddings to effectively capture distinctive environmental patterns, spatial correlations between images and cells, and spatial correlations among the images themselves.

Specifically, given cell $c_i$ and its corresponding street view images $\mathbf{sv}_i = \{\mathbf{sv}_{i,1}, \mathbf{sv}_{i,2}, \cdots \}$, we use ResNet as the image encoder to extract the initial visual embedding $\mathbf{u}_{i,j}$ for each street view image: 
{\small
\begin{equation}
    \mathbf{u}_{i,j} = \mathrm{ResNet} (\mathbf{sv}_{i,j}).
\end{equation}
}

Next, we average the visual embeddings for all street view images of the same cell $c_i$, representing the environmental context embedding of $c_i$, denoted as $\mathbf{v}_i$:
{\small
\begin{equation}
    \mathbf{v}_i = \frac{1}{|\mathbf{sv}_i|} \sum^{|\mathbf{sv}_i|}_{j=1} \mathbf{u}_{i,j},
\end{equation}
}where $|\mathbf{sv}_i|$ is the number of images associated with $c_i$.

Then, we adopt the InfoNCE~\cite{infoNCE} loss as the objective function to optimize the environment context-based image encoder. For a target street view image, the environmental context of the cell it belongs to serves as the positive sample, while the environmental contexts of other grid cells are treated as negative samples.
The objective function is defined as follows:

{\small
\begin{equation}
    \mathcal{L}^{sv} = - \frac{1}{m} \sum^{m}_{i} \sum^{\mathbf{sv}_i}_{j=i} \log 
    \left( \frac{\mathrm{exp} \left( \frac{\mathbf{u}_{i,j}^{T} \mathbf{v}_i}{\tau} \right)} 
    {\sum_{k}^{m} \mathrm{exp} \left( \frac{\mathbf{u}_{i,j}^{T} \mathbf{v}_k}{\tau}  \right) } \right).
\end{equation}
}Recall that $m$ is the total number of cells, and $\tau$ is a temperature parameter set to 0.5 in our experiment. We train the environment context-based image encoder by minimizing  $\mathcal{L}^{sv}$  to generate the street view image embeddings.

\textbf{(2) Street view-region embedding alignment.}
After obtaining the street view image embeddings, we reassign them to their corresponding regions based on the geographical locations of the images.  To handle the variability in the number of street view images per region, which can affect the subsequent alignment process, we standardize the data by randomly selecting a fixed number, $x$, of images for each region ($x$ = 64  in our experiments). 
For a given region  $r_i$,  the corresponding street view image embeddings are organized into a matrix $\mathbf{U}_i \in \mathbb{R}^{x \times d_{img}}$, where $d_{img}$ = 768.

To effectively integrate the street view image embeddings with the region embeddings learned earlier, we introduce a \emph{street view-region alignment} (SV-RAlign) module. This module extracts task-relevant ground-level visual features from the street view image embeddings, with guidance given by a downstream task.

To enable the adaptive selection of relavant visual information, we employ a cross-attention layer. 
Given region embedding $\mathbf{h}_i$ and street view image embedding $\mathbf{U}_i$ of region $r_i$, we define the query matrix $\mathbf{Q}_i = \mathbf{h}_i\mathbf{W}_Q$, key matrix $\mathbf{K}_i = \mathbf{U}_i\mathbf{W}_K$, and value matrix $\mathbf{V}_i = \mathbf{U}_i\mathbf{W}_V$, where $\mathbf{W}_Q \in \mathbb{R}^{d \times d_{proj}}$, $\mathbf{W}_K$ and $\mathbf{W}_V \in \mathbb{R}^{d_{img} \times d_{proj}}$, and $d_{proj}$ denotes the dimension of the projected matrices in latent spaces, set to 256 in our experiments. Then, we use the cross-attention operation followed by an MLP to generate the output embedding for  region $r_i$. Formally, the street view-region embedding $\mathbf{h}^{sv}_{i} \in \mathbf{H}^{sv}$ is computed as:
{\small
\begin{equation}
    \mathbf{h}_{i}^{sv} = \mathrm{MLP} \left( \mathrm{Softmax} 
    \left( \frac{\mathbf{Q}_i {\mathbf{K}_i}^T}{\sqrt{d_{proj}}}    \right) \mathbf{V}_i  \right),
\end{equation}
}

\subsubsection{Model Training.}
\label{subsubsec:modeltrain_downstream}
After obtaining the region embedding $\mathbf{H}$, the text-region embedding $\mathbf{H}^{t}$, and the street view-region embedding $\mathbf{H}^{sv}$, we construct the final regin embeddings $\hat{\mathbf{h}_i} \in \hat{\mathbf{H}}$ as $ \hat{\mathbf{h}_i} = \mathbf{h}_i \; || \; \mathbf{h}^{t}_{i} \; || \; \mathbf{h}^{sv}_{i}$,
where `$||$' denotes concatenation. We use a feedforward neural network (FNN) for a given downstream (prediction) task, formulated as $ \hat{y_i} = \mathrm{FFN} (\hat{\mathbf{h}_i})$, 
where $\hat{y_i}$ is the prediction output for region $r_i$. To optimize the \prompt\ module, we adopt the mean squared error loss:
{\small
\begin{equation}
    \mathcal{L}_{pe} = \frac{1}{n} \sum_{i=1}^{n} \left(
        \hat{y_i} - y_i
    \right)^2,
\end{equation}
}where $y_i$ is the ground truth for $r_i$, and $n$ is the  number of regions.

\section{Experiments}
\label{sec:exp}

We run experiments to verify: 
(\textbf{Q1})~the embedding quality of our \model\ model as compared with the state-of-the-art (SOTA) models on four downstream tasks,
(\textbf{Q2})~the applicability of \model\ across diverse geographic regions,
(\textbf{Q3})~the adaptability of our cell embeddings to different region formations,
(\textbf{Q4})~the impact of our model components and  input features, 
(\textbf{Q5})~the applicability of \model\ to areas of different urban environments
(\textbf{Q6})~the impact of the grid cell design, prompt templates, and key hyper-parameters.

\subsection{Experimental Settings}
\label{subsec:exp_settings}

\textbf{Dataset.}
We use real data from five cities across the globe: New York City (\textbf{NYC})~\cite{nycOpendata}, Chicago (\textbf{CHI})~\cite{chiOpendata}, San Francisco (\textbf{SF})~\cite{sfOpendata}, Singapore City (\textbf{SG})~\cite{sgOpendata}, and Lisbon (\textbf{LX})~\cite{lxRegion}. We collect data on region division, POI, land use, satellite images, street view images, check-in, and population.  Additionally, crime and service call data are collected for NYC, CHI, and SF (these are unavailable for the other two cities).

\begin{table}[ht]
\centering
\caption{Dataset Statistics (New York City, Chicago, and San Francisco)}
\label{tab:datasets}
\setlength{\tabcolsep}{2pt}
\resizebox{\columnwidth}{!}{
\begin{tabular}{lrrr}
\toprule
&\textbf{NYC~\cite{nycOpendata}} & \textbf{CHI~\cite{chiOpendata}} & \textbf{SF~\cite{sfOpendata}} \\ \midrule
\#regions & 180 & 77 & 175 \\

\#grid cells & 438 & 720 & 1032 \\

\#POIs & 24,496 & 57,891 & 28,578 \\

\#POI categories & 15 & 15 & 15 \\

\#land use categories  & 20 & 20 & 20 \\

\#satellite images  & 438 & 720 & 1,032 \\

\#street view images  & 29,336 & 61,154 & 43,854 \\

{\#crime records} & 35,335 & 18,200  & 48,489 \\
(data collection time) & unknown & 12/2022 - 12/2022 & 01/2022 - 12/2022 \\

{\#check-ins} & 106,902 & 167,232 & 87,750 \\
(data collection time) & 04/2012 - 09/2013 & 04/2012 - 09/2013 & 04/2012 - 09/2013 \\

{\#service calls} & 516,187 & 24,350 & 34,385 \\ 
(data collection time) & 01/2023 - 03/2023 & 12/2022 - 12/2022 & 01/2022 - 12/2022 \\ 

Population counts & 1,540,692 & 2,508,984 & 801,251 \\
(data collection time) & 2020 & 2020 & 2020 \\

\bottomrule
\end{tabular}
}
\vspace{-1mm}
\end{table}

Table~\ref{tab:datasets} summarizes the  NYC, CHI, and SF datasets. Each dataset consists of region boundaries obtained from open data portals, POIs extracted from OpenStreetMap~\cite{osm} with category labels detailed in Section~\ref{subsubsec:feature_collection}, and land use data also sourced from OpenStreetMap. Satellite images are collected from Google Maps~\cite{googlemap} at a fixed resolution of 800×800 pixels, while street view images have a resolution of 640×500 pixels and  cover sampled points distributed across the full area of interest (the sampling strategy is detailed in Section~\ref{subsubsec:feature_collection}). Crime and service call records are retrieved from open data portals, while check-in records are obtained from a Foursquare dataset~\cite{checkinData}. Each record contains location and time information, with counts aggregated at region level. Population data for 2020 is sourced from WorldPop~\cite{WorldPop}.

\begin{table}[ht]
\centering
\caption{Dataset Statistics (Singapore and Lisbon)}
\label{tab:datasets_2}
\setlength{\tabcolsep}{6pt}
\resizebox{0.9\columnwidth}{!}{
\begin{tabular}{lrr}
\toprule
&\textbf{SG~\cite{sgOpendata}} & \textbf{LX~\cite{lxRegion}}
\\ \midrule
\#regions & 324 & 53  \\

\#spatial partition units & 748 & 690  \\

\#POIs & 65,082 & 43,961  \\

\#POI categories & 15 & 15  \\

\#land use categories  & 20 & 20  \\

\#satellite images  & 748 & 690  \\

\#street view images  & 56,102 & 39,656  \\

{\#check-ins} & 355,463 & 24,327  \\
(data collection time) & 04/2012 - 09/2013 & 04/2012 - 09/2013 \\

Population counts & 4,296,918 & 507,846 \\
(data collection time) & 2020 & 2020 \\

\bottomrule
\end{tabular}
}
\vspace{-1mm}
\end{table}

Table~\ref{tab:datasets_2} summarizes the SG and LX datasets,  which contain the same data features as above except for crime and service call data. 

\textbf{Competitors.}
We compare with models from two categories. The first category uses a subset of the publicly accessible data: \textbf{RegionDCL}~\cite{RegionDCL}, \textbf{UrbanCLIP}~\cite{urbanclip}, and \textbf{CityFM}~\cite{cityFM} (SOTA). The second category uses human mobility data, which has restricted availability: \textbf{MVURE}~\cite{MVURE}, \textbf{MGFN}~\cite{MGFN}, \textbf{HREP}~\cite{HREP}, \textbf{ReCP}~\cite{ReCP}, and \textbf{HAFusion}~\cite{HAFusion} (SOTA). This latter category does not apply to SG and LX due to lack of  data. 

Models based on readily accessible data: 
\begin{itemize}
\item {\textbf{RegionDCL~\cite{RegionDCL}}} partitions the buildings within a region into non-overlapping groups. It then computes the embeddings of these building groups through contrastive learning, both within the group (between the group and the buildings inside) and between the group and its corresponding region. Finally, the embeddings of the building groups within a region are aggregated to generate the region embedding.
\item {\textbf{UrbanCLIP~\cite{urbanclip}}} generates detailed textual descriptions for each satellite image corresponding to a region, forming image-text pairs. Subsequently, it is trained on the these pairs using contrastive learning to generate text-enhanced visual representations of the satellite images, which serve as the embeddings for their associated regions.
\item {\textbf{CityFM~\cite{cityFM}}} (SOTA)  
leverages geospatial entities (e.g., buildings and road segments) extracted from OpenStreetMap and employs contrastive learning with three objectives to generate entity embeddings: a mutual information-based text-to-text objective, a vision-language objective, and a road-based context-to-context objective. The resulting entity embeddings are then aggregated to generate corresponding region embeddings.
\end{itemize}

Models based on human mobility data: 
\begin{itemize}
\item {\textbf{MVURE~\cite{MVURE}}} constructs multiple graphs with regions as vertices, using human mobility, POI, and check-in features. It then applies GAT to each graph to learn embeddings. Finally, it generates the final region embeddings by performing a weighted summation of the embeddings from each graph. 
\item {\textbf{MGFN~\cite{MGFN}}} 
constructs multiple mobility graphs based on the human mobility features.  It then clusters these graphs to form mobility pattern graphs based on their spatio-temporal distances. Finally, message passing is performed on the mobility pattern graphs to generate the region embeddings.
\item {\textbf{HREP~\cite{HREP}}} uses human mobility, POI, and geographic neighbour features to generate region embeddings. Subsequently, it randomly generates learnable prompt embeddings and concatenates them with the region embeddings, tailoring the embeddings for different downstream tasks.
\item {\textbf{ReCP~\cite{ReCP}}} uses human mobility and POI features to generate region embeddings via a multi-view learning approach.
Instead of fusing multi-view information in a posterior stage, it does the integration by employing two objective functions: maximizing mutual information between views and minimizing conditional entropy.
\item {\textbf{HAFusion~\cite{HAFusion}}} (SOTA)  uses human mobility, POI, and land use features to generate region embeddings. It employs an attention-based fusion module to fuse multi-view information both at the region level and the view level, to capture higher-order correlations among the regions.
\end{itemize}




\textbf{Model hyperparameter settings.}
All models were trained and tested on a machine equipped with an NVIDIA Tesla V100 GPU and 64 GB of memory.

For the competitor models, we follow parameter settings recommended in their papers as much as possible. We use special settings as described in the HAFusion paper~\cite{HAFusion} that reduce the model scales on CHI for MGFN, MVURE, HREP, and ReCP, as  this dataset has fewer regions. The same applies to LX as it has fewer regions as well. 
RegionDCL, UrbanCLIP, and CityFM do not require special settings, as their training processes are determined by the number of building groups, image-text pairs, and geospatial entities, respectively, rather than the number of regions. 

For our grid cell embedding learning model, the number of layers in the GNN branch of the grid-based intra-view feature learning module is 3 on New York City, and 2 for the other datasets. In the grid-based dual-feature attentive fusion module, the number of layers is 3 for all datasets. We train it for 2,000 epochs in full batches, using Adam optimization with a learning rate of 0.0001. For our downstream task learning model, the dimensionality of the text-region embeddings is 144, and the number of street view images used in the street view-region alignment module is 64. We train this model for 1,000 epochs in full batches, using Adam optimization with a learning rate of 0.0005 and weight decay of 0.0005. These hyperparameter values are set by a grid search.

The region embedding dimensionality $d$ is set as 144 for our model following HAFusion and HREP.
The region embedding dimensionalities for MVURE, MGFN, RegionDCL, and ReCP are 96, 96, 64, and 96, respectively, as suggested by their original papers. Our experimental results in Section~\ref{subsubsec:impact_of_region_dim} also show that these dimensionality values are optimal for the respective models (i.e., their yielded embeddings are of lower quality when $d$ is 144). 
The region embedding dimensionalities of UrbanCLIP and CityFM are difficult to change from their default implementation. For UrbanCLIP, $d = 768$ which is determined by the image encoder, as it uses the  embeddings of satellite images from a vision language model CLIP~\cite{CLIP} as the  region embedding. Similarly, for CityFM, $d = 1792$ which is determined by the dimensionality of the embeddings of different geospatial entities enclosed by a region. 

As noted in the HAFusion paper~\cite{HAFusion}, MVURE takes check-in records as part of its input. This model is used for the check-in prediction task regardless. We use data from non-overlapping time periods for the training and testing processes of the model, such that it does not see the testing data at training.



\textbf{Evaluation procedure.} We use each representation learning model to generate region embeddings for each city separately. The embeddings then serve as input to machine learning models for downstream tasks (i.e., downstream models). We use four downstream prediction tasks following the baseline models~\cite{HAFusion, RegionDCL}: crime, check-in, service call, and population counts.  Since these tasks are regression-based, we employ a ridge regression model for each task, with ten-fold cross-validation. 

We evaluate the representation learning models through the downstream models in mean absolute error (\textbf{MAE}), root mean square error (\textbf{RMSE}), and coefficient of determination ($\boldsymbol{R^2}$).

\subsection{Overall Results (Q1)}
\label{sub:overall_results}

The overall model accuracy results are reported in Table~\ref{tab:overall_results_full_ver}. We make the following observations.


\begin{table*}[htbp] \scriptsize
\caption{\small Overall Prediction Accuracy Results (`$\downarrow$' indicates that smaller values are preferred, and `$\uparrow$' indicates that large values are preferred. The best results are in boldface, and the second-best results are underlined.)}
\label{tab:overall_results_full_ver}
\setlength{\tabcolsep}{1.5pt} 
\resizebox{\textwidth }{!}{
\begin{tabular}{l |c c c|c c c|c c c| c c c }
    \toprule [0.4ex] \\[-3.5ex]

    \multirow{2}{*}{\textbf{\makecell[l]{New York \\ City}}}& \multicolumn{3}{c|}{Crime} & \multicolumn{3}{c|}{Check-in} & \multicolumn{3}{c|}{Service Call} & \multicolumn{3}{c}{Population} \\
    \cmidrule(lr){2-4} \cmidrule(lr){5-7} \cmidrule(lr){8-10}  \cmidrule(lr){11-13}
     &MAE $\downarrow$ & RMSE $\downarrow$ & $R^{2} \uparrow$ 
     & MAE $\downarrow$ & RMSE $\downarrow$ & $R^{2} \uparrow$ 
     & MAE $\downarrow$ & RMSE $\downarrow$ & $R^{2} \uparrow$ 
     & MAE $\downarrow$ & RMSE $\downarrow$ & $R^{2} \uparrow$ \\
    \midrule
    MVURE~\cite{MVURE} & 67.9 $\pm$ 1.1 & \, 93.8 $\pm$ 1.9 & 0.591 $\pm$ 0.016 & 
    306.7 $\pm$ 8.20 & 499.6 $\pm$ 12.9 & 0.627 $\pm$ 0.019 &
    1428 $\pm$ 33 & 2180 $\pm$ \, 46 & 0.367 $\pm$ 0.027 &
    2814 $\pm$ 67 & 3564 $\pm$ \, 79 & 0.545 $\pm$ 0.008 
    \\
    MGFN~\cite{MGFN} & 70.2 $\pm$ 2.3 & \, 89.6 $\pm$ 2.5 & 0.630 $\pm$ 0.020  & 
    292.6 $\pm$ 17.1 & 451.8 $\pm$ 28.1 & 0.690 $\pm$ 0.040  &
    1554 $\pm$ 81 & 2286 $\pm$ 115 & 0.303 $\pm$ 0.069  &
    2822 $\pm$ 42 & 3706 $\pm$ \, 98 & 0.509 $\pm$ 0.026  
    \\
    HREP~\cite{HREP} & 62.8 $\pm$ 2.1 & \, 83.1 $\pm$ 2.3 & 0.680 $\pm$ 0.014  & 
    276.3 $\pm$ 11.7 & 448.2 $\pm$ 17.1 & 0.703 $\pm$ 0.021  &
    1430 $\pm$ 29 & 2286 $\pm$ \, 34 & 0.398 $\pm$ 0.021 &
    2656 $\pm$ 59 & 3461 $\pm$ \, 83 & 0.571 $\pm$ 0.021 
    \\
    ReCP~\cite{ReCP} & 83.1 $\pm$ 2.4 & 108.7 $\pm$ 2.2 & 0.459 $\pm$ 0.022 & 
    246.9 $\pm$ \, 4.3 & 400.1 $\pm$ 23.6 & 0.761 $\pm$ 0.029 &
    1516 $\pm$ 66 & 2199 $\pm$ \, 49 & 0.356 $\pm$ 0.029 & 
    3305 $\pm$ 90 & 4353 $\pm$ 153 & 0.322 $\pm$ 0.047 
    \\
    HAFusion~\cite{HAFusion} & \underline{56.1 $\pm$ 1.3} & \underline{\, 76.1 $\pm$ 2.2} & \underline{0.734 $\pm$ 0.015} & 
    \underline{202.8 $\pm$ \, 7.2} & \underline{322.8 $\pm$ 12.6} & \underline{0.844 $\pm$ 0.012} & 
    \underline{1273 $\pm$ 20} & \underline{1951 $\pm$ \, 27} & \underline{0.493 $\pm$ 0.014} & 
    \underline{2497 $\pm$ 50} & \underline{3277 $\pm$ \, 82} & \underline{0.616 $\pm$ 0.019} 
    \\
    \midrule 
    RegionDCL~\cite{RegionDCL} & 98.7 $\pm$ 3.1 & 127.9 $\pm$ 5.2 & 0.251 $\pm$ 0.026  &
    371.2 $\pm$ 10.3 & 495.5 $\pm$ 15.9 & 0.471 $\pm$ 0.023  &
    1783 $\pm$ 21 & 2597 $\pm$ \, 38 & 0.103 $\pm$ 0.026  &
    3753 $\pm$ 47 & 4734 $\pm$ \, 59 & 0.198 $\pm$ 0.019 
    \\
    UrbanCLIP~\cite{urbanclip} & 97.4 $\pm$ 2.6 & 126.1 $\pm$ 1.9 & 0.267 $\pm$ 0.012  &
    393.6 $\pm$ \, 5.9 & 602.4 $\pm$ \, 3.1 & 0.458 $\pm$ 0.005  &
    1409 $\pm$ \, 7 & 2401 $\pm$ \, 16 & 0.232 $\pm$ 0.005 &
    3338 $\pm$ 11 & 4499 $\pm$ \, 16 & 0.276 $\pm$ 0.002 
    \\
    CityFM~\cite{cityFM} & 95.5 $\pm$ 1.4 & 122.4 $\pm$ 1.8 & 0.315 $\pm$ 0.010  & 
    380.2 $\pm$ \, 3.8 & 594.9 $\pm$ \, 6.4 & 0.471 $\pm$ 0.011  &
    1781 $\pm$ 28 & 2578 $\pm$ \, 19 & 0.117 $\pm$ 0.013  &
    3515 $\pm$ 18 & 4545 $\pm$ \, 26 & 0.261 $\pm$ 0.002  
    \\
    \midrule 
    \textbf{\model} & \textbf{50.4 $\pm$ 1.1} & \textbf{\, 67.6 $\pm$ 1.5} & \textbf{0.789 $\pm$ 0.009}  & 
    \textbf{187.3 $\pm$ \, 5.0} & \textbf{287.5 $\pm$ \, 7.6} & \textbf{0.876 $\pm$ 0.006}  &
    \textbf{1131 $\pm$ 47} & \textbf{1727 $\pm$ \, 46} & \textbf{0.601 $\pm$ 0.021}  &
    \textbf{2159 $\pm$ 28} & \textbf{2822 $\pm$ \, 47} & \textbf{0.715 $\pm$ 0.010}  
    \\
    \midrule 
    \textbf{Improvement} & \textbf{10.2\%} & \textbf{11.2\%} &\textbf{7.5\%} &
    \textbf{7.6\%} &\textbf{10.9\%} &\textbf{3.8\%} &
    \textbf{11.2\%} &\textbf{11.5\%} &\textbf{21.9\%} &
    \textbf{13.5\%} &\textbf{13.9\%} &\textbf{16.1\%} 
    \\
    \bottomrule \toprule [0.4ex] \\[-3.5ex]

    \multirow{2}{*}{\textbf{\makecell[l]{Chicago}}}& \multicolumn{3}{c|}{Crime} & \multicolumn{3}{c|}{Check-in} & \multicolumn{3}{c|}{Service Call} & \multicolumn{3}{c}{Population}\\
    \cmidrule(lr){2-4} \cmidrule(lr){5-7} \cmidrule(lr){8-10}
     \cmidrule(lr){2-4} \cmidrule(lr){5-7} \cmidrule(lr){8-10}  \cmidrule(lr){11-13}
     &MAE $\downarrow$ & RMSE $\downarrow$ & $R^{2} \uparrow$ 
     & MAE $\downarrow$ & RMSE $\downarrow$ & $R^{2} \uparrow$ 
     & MAE $\downarrow$ & RMSE $\downarrow$ & $R^{2} \uparrow$ 
     & MAE $\downarrow$ & RMSE $\downarrow$ & $R^{2} \uparrow$ \\
    \midrule
    MVURE~\cite{MVURE} & 100.4 $\pm$ 6.6 & 129.2 $\pm$ 7.3 & 0.461 $\pm$ 0.062  & 
    1693 $\pm$ \, 74 & 3171 $\pm$ 128 & 0.656 $\pm$ 0.029  & 
    190.3 $\pm$ \, 9.8 & 266.9 $\pm$ 12.1 & 0.441 $\pm$ 0.050  &
    13717 $\pm$ 322 & 17174 $\pm$ 552 & 0.313 $\pm$ 0.043 
    \\
    MGFN~\cite{MGFN} & 107.4 $\pm$ 5.4 & 137.9 $\pm$ 5.2 & 0.386 $\pm$ 0.047  &
    1281 $\pm$ \, 41 & 2276 $\pm$ \, 86 & 0.817 $\pm$ 0.011  &
    208.2 $\pm$ 11.3 & 293.4 $\pm$ 16.6 & 0.329 $\pm$ 0.077  &
    13071 $\pm$ 505 & 16578 $\pm$ 707 & 0.359 $\pm$ 0.054 
    \\
    HREP~\cite{HREP} & \, 88.3 $\pm$ 6.4 & 114.4 $\pm$ 5.5 & 0.578 $\pm$ 0.041  & 
    1679 $\pm$ \, 71 & 3135 $\pm$ \, 79 & 0.664 $\pm$ 0.017  &
    185.7 $\pm$ \, 6.1 & 262.2 $\pm$ 10.8 & 0.468 $\pm$ 0.022  &
    12063 $\pm$ 539 & 15397 $\pm$ 832 & 0.447 $\pm$ 0.061  
    \\
    ReCP~\cite{ReCP} & \, 86.9 $\pm$ 5.5 & 120.1 $\pm$ 7.1 & 0.534 $\pm$ 0.057  &
    1272 $\pm$ \, 92 & 2341 $\pm$ 267 & 0.804 $\pm$ 0.045  &
    206.7 $\pm$ 11.1 & 303.4 $\pm$ 16.1 & 0.284 $\pm$ 0.076 & 
    12085 $\pm$ 400 & 17029 $\pm$ 561 & 0.325 $\pm$ 0.044 
    \\
    HAFusion~\cite{HAFusion} & \underline{\, 77.8 $\pm$ 3.6} & \underline{107.1 $\pm$ 5.4} & \underline{0.631 $\pm$ 0.036} & 
    \underline{\, 929 $\pm$ \, 62} & \underline{1947 $\pm$ \, 75} & \underline{0.870 $\pm$ 0.010} &
    \underline{159.3 $\pm$ 13.9} & \underline{222.0 $\pm$ 18.9} & \underline{0.613 $\pm$ 0.067} &
    \underline{10678 $\pm$ 390} & \underline{13988 $\pm$ 548} & \underline{0.544 $\pm$ 0.035} 
    \\
    \midrule 
    RegionDCL~\cite{RegionDCL} & 121.7 $\pm$ 4.8 & 159.6 $\pm$ 6.3 & 0.179 $\pm$ 0.053  &
    2427 $\pm$ 123 & 4184 $\pm$ 136 & 0.402 $\pm$ 0.042  &
    195.7 $\pm$ \, 7.6 & 272.1 $\pm$ 10.1 & 0.445 $\pm$ 0.041  &
    14289 $\pm$ 343 & 18653 $\pm$ 368 & 0.190 $\pm$ 0.032 
    \\
    UrbanCLIP~\cite{urbanclip} & 101.6 $\pm$ 0.6 & 134.7 $\pm$ 1.7 & 0.416 $\pm$ 0.006  &
    2612 $\pm$ \, 29 & 4885 $\pm$ \, 73 & 0.186 $\pm$ 0.024  &
    183.2 $\pm$ \, 0.9 & 256.3 $\pm$ \, 1.8 & 0.491 $\pm$ 0.003  &
    13328 $\pm$ \, 69 & 17498 $\pm$ \, 74 & 0.288 $\pm$ 0.006  
    \\
    CityFM~\cite{cityFM} & 121.6 $\pm$ 1.8 & 157.1 $\pm$ 2.8 & 0.205 $\pm$ 0.018  & 
    1980 $\pm$ \, 64 & 3362 $\pm$ 109 & 0.614 $\pm$ 0.025  &
    198.3 $\pm$ \, 3.7 & 280.1 $\pm$ \, 6.1 & 0.391 $\pm$ 0.027 &
    13904 $\pm$ \, 37 & 17704 $\pm$ \, 56 & 0.271 $\pm$ 0.004
    \\
    \midrule 
    \textbf{\model} & \textbf{\, 61.7 $\pm$ 3.5} & \textbf{\, 85.1 $\pm$ 4.2} & \textbf{0.766 $\pm$ 0.022}  & 
    \textbf{\,  922 $\pm$ \, 76} & \textbf{1775 $\pm$ \, 199} & \textbf{0.891 $\pm$ 0.024}  &
    \textbf{121.1 $\pm$ \, 7.4} & \textbf{178.2 $\pm$ \, 9.5} & \textbf{0.753 $\pm$ 0.026}  &
    \textbf{8126 $\pm$ 224} & \textbf{11395 $\pm$ 255} & \textbf{0.698 $\pm$ 0.014}
    \\
    \midrule 
    \textbf{Improvement} & \textbf{20.7\%} &\textbf{20.5\%} & \textbf{21.4\%}  &
    \textbf{0.7\%} &\textbf{8.8\%} &\textbf{2.4\%} &
    \textbf{24.0\%} &\textbf{19.7\%} &\textbf{22.8\%} &
    \textbf{23.9\%} &\textbf{18.5\%} &\textbf{28.3\%}
    \\
    \bottomrule \toprule [0.4ex] \\[-3.5ex]

    \multirow{2}{*}{\textbf{\makecell[l]{San \\ Francisco}}}& \multicolumn{3}{c|}{Crime} & \multicolumn{3}{c|}{Check-in} & \multicolumn{3}{c|}{Service Call} & \multicolumn{3}{c}{Population}\\
    \cmidrule(lr){2-4} \cmidrule(lr){5-7} \cmidrule(lr){8-10}
     \cmidrule(lr){2-4} \cmidrule(lr){5-7} \cmidrule(lr){8-10}  \cmidrule(lr){11-13}
     &MAE $\downarrow$ & RMSE $\downarrow$ & $R^{2} \uparrow$ 
     & MAE $\downarrow$ & RMSE $\downarrow$ & $R^{2} \uparrow$ 
     & MAE $\downarrow$ & RMSE $\downarrow$ & $R^{2} \uparrow$ 
     & MAE $\downarrow$ & RMSE $\downarrow$ & $R^{2} \uparrow$ \\
    \midrule
    MVURE~\cite{MVURE} & 130.3 $\pm$ 1.7 & 201.7 $\pm$ \, 3.2 & 0.594 $\pm$ 0.013  
    & 346.8 $\pm$ \, 8.7 & 659.3 $\pm$ 15.7 & 0.562 $\pm$ 0.021  
    & 102.1 $\pm$ 4.8 & 164.7 $\pm$ \, 2.7 & 0.479 $\pm$ 0.017  
    & 1466 $\pm$ 20 & 1901 $\pm$ 27 & \, 0.093 $\pm$ 0.016 
    \\
    MGFN~\cite{MGFN} & 128.4 $\pm$ 3.3 & 199.9 $\pm$ \, 4.3 & 0.601 $\pm$ 0.017  
    & 310.8 $\pm$ \, 9.1 & 542.1 $\pm$ 17.6 & 0.708 $\pm$ 0.010  
    & 102.8 $\pm$ 2.2 & 166.3 $\pm$ \, 2.5 & 0.468 $\pm$ 0.021 
    & 1527 $\pm$ 39 & 1964 $\pm$ 49 & \, 0.033 $\pm$ 0.048  
    \\
    HREP~\cite{HREP} & 124.4 $\pm$ 2.3 & 196.9 $\pm$ \, 3.9 & 0.612 $\pm$ 0.014 
    & 330.9 $\pm$ \, 9.3 & 606.7 $\pm$ 25.8 & 0.629 $\pm$ 0.032 
    & 103.4 $\pm$ 3.2 & 167.4 $\pm$ \, 4.6 & 0.461 $\pm$ 0.029 
    & 1436 $\pm$ 40 & 1867 $\pm$ 45 & \, 0.127 $\pm$ 0.023 
    \\
    ReCP~\cite{ReCP} & 115.4 $\pm$ 7.9 & 202.9 $\pm$ 18.6 & 0.585 $\pm$ 0.075  
    & 233.9 $\pm$ 12.5 & 462.2 $\pm$ 30.9 & 0.783 $\pm$ 0.029  
    & 108.5 $\pm$ 6.4 & 190.2 $\pm$ 16.8 & 0.301 $\pm$ 0.119 
    & 1471 $\pm$ 49 & 1929 $\pm$ 46 & \, 0.067 $\pm$ 0.044
    \\
    HAFusion~\cite{HAFusion} & \underline{101.5 $\pm$ 3.3} & \underline{178.4 $\pm$ \, 3.6} & \underline{0.682 $\pm$ 0.013}  
    & \underline{233.1 $\pm$ \, 9.5} & \underline{429.6 $\pm$ 28.1} & \underline{0.813 $\pm$ 0.024}  
    & \underline{\, 81.5 $\pm$ 2.5} & \underline{142.1 $\pm$ \,  3.2} & \underline{0.612 $\pm$ 0.018}  
    & \underline{1387 $\pm$ 28} & \underline{1833 $\pm$ 17} & \underline{\, 0.159 $\pm$ 0.006}
    \\
    \midrule 
    RegionDCL~\cite{RegionDCL} & 156.3 $\pm$ 2.1 & 242.3 $\pm$ 4.6 & 0.413 $\pm$ 0.021 
    & 398.8 $\pm$ \, 9.9 & 748.1 $\pm$ 17.8 & 0.437 $\pm$ 0.024 
    & 116.6 $\pm$ 2.3 & 196.7 $\pm$ \, 3.2 & 0.256 $\pm$ 0.024 
    & 1513 $\pm$ 32 & 1971 $\pm$ 25 & \, 0.027 $\pm$ 0.025
    \\
    UrbanCLIP~\cite{urbanclip} & 171.1 $\pm$ 1.0 & 269.8 $\pm$ 2.6 & 0.283 $\pm$ 0.014 
    & 380.3 $\pm$ \, 2.6 & 813.5 $\pm$ \, 3.3 & 0.334 $\pm$ 0.002 
    & 106.9 $\pm$ 1.1 & 192.1 $\pm$ \, 1.2 & 0.292 $\pm$ 0.008 
    & 1695 $\pm$ 17 & 2360 $\pm$ 34 & -0.395 $\pm$ 0.005
    \\
    CityFM~\cite{cityFM} & 168.3 $\pm$ 0.6 & 259.8 $\pm$ 1.5 & 0.334 $\pm$ 0.008 
    & 428.3 $\pm$ \, 2.7 & 839.3 $\pm$ \, 4.7 & 0.298 $\pm$ 0.008 
    & 105.1 $\pm$ 1.3 & 178.5 $\pm$ \, 1.6 & 0.396 $\pm$ 0.008 
    & 1578 $\pm$ 14 & 1982 $\pm$ 28 & \, 0.023 $\pm$ 0.003
    \\
    \midrule 
    \textbf{\model} & \textbf{\, 98.6 $\pm$ 3.9} & \textbf{163.7 $\pm$ 4.3} & \textbf{0.732 $\pm$ 0.014} 
    & \textbf{229.4 $\pm$ \, 8.1} & \textbf{375.2 $\pm$ 34.9} & \textbf{0.859 $\pm$ 0.011} 
    & \textbf{\, 79.9 $\pm$ 4.6} & \textbf{136.5 $\pm$ \, 3.8} & \textbf{0.641 $\pm$ 0.021}
    & \textbf{1032 $\pm$ 29} & \textbf{1441 $\pm$ 46} & \textbf{\, 0.480 $\pm$ 0.034} 
    \\
    \midrule 
    \textbf{Improvement} & \textbf{2.9\%} &\textbf{8.2\%} &\textbf{7.3\%} 
    & \textbf{1.7\%} & \textbf{12.7\%} & \textbf{5.7\%} 
    & \textbf{2.0\%} & \textbf{3.9\%} & \textbf{4.7\%} 
    &\textbf{25.6\%} &\textbf{21.4\%} &\textbf{202\%} 
    \\

    \bottomrule
\end{tabular}
}
\end{table*}



(1)~Our model \model\ outperforms all competitors including even those using human mobility data in addition, across three cities in the USA (the other two cities will be shown next) and all four downstream tasks, improving $R^2$ by up to 202\% over the best baseline HAFusion.
This is attributed to our novel model design: (i) Our grid cell-based embeddings and their adaptive aggregation help capture local variations within regions, ensuring that the learned embeddings accurately reflect nuanced region characteristics. (ii)~Our prompt enhanced embeddings extract task-specific information to meet the specific needs of different tasks. 
\model\ excels particularly on population prediction, as the street view images provide information such as  building density and types, which strongly correlate with population distribution.

(2)~The baseline models using human mobility data (e.g., HAFusion) outperform those based on publicly accessible data (e.g., CityFM) for  most datasets and downstream tasks. This is because human mobility reflects population distribution and movement patterns of individuals, which are closely related to the downstream tasks, especially check-in count prediction for which these models perform particularly well. 
This highlights the difficulties and our technical contributions in designing a model that outperforms the mobility data-based models without using mobility data.

(3)~The baseline models using readily accessible data perform poorly for the following reasons. RegionDCL  uses only building footprints. It struggles to distinguish the different functionality of regions and hence their crime, check-in, service call, and population counts. 
UrbanCLIP uses satellite images and their textual descriptions generated by a vision language model (VLM). Satellite images are coarse-grained. Meanwhile, VLMs are prone to generating low-quality textual descriptions due to hallucination, which limits the capability of UrbanCLIP.
CityFM was designed to generate embeddings for different geospatial entities (e.g., buildings and roads). Simply concatenating these entity embeddings to form region embeddings can dilute the unique region characteristics.

\textbf{Model running time.}
Table~\ref{tab:computation_time} reports the running times for embedding learning and downstream task learning (we omit the times on SG and LX as the comparative patterns resemble). The downstream task running times include both model training (prompt learning) and inference, with inference times in parentheses. The downstream tasks share identical input and output sizes.

Our model \model\ requires additional time to learn the region embeddings because it starts with learning embeddings for the cells, and there are more cells than regions. Additionally, our model takes additional time for downstream task training, as it integrates textual features and street view visual features into the region embeddings through two alignment modules. Adding these times together, our embedding learning can still be done in less than 10 minutes, which is rather affordable for training a model with deep learning. While \model\ takes extra learning times, it significantly reduces prediction errors, as shown in the experiments above.

\begin{table}[ht]
\centering
\caption{Embedding Learning and Testing Times (seconds)}
\label{tab:computation_time}
\renewcommand{\arraystretch}{1.1}
\resizebox{1\columnwidth}{!}{
\begin{tabular}{l|rrr|rrr}
\toprule

&\multicolumn{3}{c|}{\textbf{Embedding Learning}} & \multicolumn{3}{c}{\textbf{Downstream Task}} \\ \hline 
& \textbf{NYC} & \textbf{CHI} & \textbf{SF} & \multicolumn{1}{c}{\textbf{NYC}} & \multicolumn{1}{c}{\textbf{CHI}} & \multicolumn{1}{c}{\textbf{SF}} \\ 
\hline
MVURE & \textbf{35} & \textbf{15} & \textbf{34} & 0.023 (0.001)& \textbf{0.053} (0.002) & 0.026 (0.001)\\ 
\hline

MGFN & 92 & 123 & 47 &0.019 (0.001)  &0 .061 (0.002) & 0.029 (0.001)\\ 
\hline

HREP& 51 & 45 & 51 &92 (0.003) &146 (0.005) &91 (0.004)\\ 
\hline

ReCP & 178 & 80 & 180 &0.020 (0.001)&0.056 (0.002)&0.028 (0.001)\\ 
\hline

HAFusion & 79 & 51 & 78 &0.022 (0.001)&0.061 (0.002)&0.028 (0.001)\\ 
\hline

RegionDCL & 149 & 1,779 & 324 & \textbf{0.017} (0.001) &0.054 (0.002) &\textbf{0.023 (0.001)}\\ 
\hline

UrbanCLIP &1,532 &2,497 &3,359 & 86 (0.005) &86 (0.005) & 84 (0.005)\\ 
\hline

CityFM &7,521 &8,265 &7,339 & 104 (0.006)&102 (0.005) &104 (0.006)\\ 
\hline

\model\ &208 &282 &436 & 137 (0.007) &103 (0.006) &142 (0.007)\\ 
\bottomrule
\end{tabular}
}
\end{table}

The models that use readily accessible data, UrbanCLIP and CityFM, take more times for embedding learning than ours. This is because of large training datasets and complex model structures. These two models also take extra time at the training stage for the downstream tasks as they need to extract useful information from embedding by using deep networks with multiple layers.
RegionDCL is slow on CHI, because the number of buildings is extremely high compared to other cities. 
The other baseline models, which use human mobility data, are typically faster in embedding learning 00, as human mobility data provides critical insights into regional interactions and the functional relationships between regions, enabling the model to converge quickly.

The inference times for the downstream tasks are similar across all models, as all prediction models share a simple structure.

\begin{table}[htbp] 
\caption{Impact of $\#$Cells to Embedding Learning Model Running Time (in seconds)}
\label{tab:model_runtime_NO_cells}
\begin{center}
\setlength{\tabcolsep}{8pt}
\resizebox{\columnwidth }{!}{
\begin{tabular}{l*{5}{c}}
    \toprule
    $\#$Cell & 128 & 256 & 512 & 1024 & 2048 \\
    \midrule
    
   Training time & 73 & 125 & 228 & 445 & 803 \\
    
    Inference time & 0.055 & 0.105 & 0.204 & 0.412 & 0.746 \\
    \bottomrule
\end{tabular}
}
\end{center}
\end{table}

We further evaluat model running times as the number of cells varies from 128 to 2048 (cells are our basic embedding unit). As shown Table~\ref{tab:model_runtime_NO_cells}, the running times increase linearly with the number of cells. With 2048 cells, training takes about 10 minutes, and inference takes less than 1 second (on an NVIDIA Tesla V100 GPU). These results validate our model scalability.

\subsection{Cross-country Applicability (Q2)}
We further show the applicability of \model\ over cities outside the USA, i.e., Singapore (Asia) and Lisbon (Europe). We compare with the models (RegionDCL, UrbanCLIP, and CityFM) using readily accessible data for the check-in and population count prediction tasks, as there is no mobility or crime/service call count data. 
\begin{table}[htbp] 
\captionsetup{justification=centering}
\caption{\small Prediction Accuracy over Cities in Different Countries}
\label{tab:model_adaptability_partial}
\vspace{-3mm}
\setlength{\tabcolsep}{3pt} 
\renewcommand{\arraystretch}{0.9} 
\resizebox{\columnwidth }{!}{
\begin{tabular}{l |c c|c c}
    \toprule [0.4ex] \\[-3ex]

    & \multicolumn{2}{c|}{\textbf{Singapore}} & \multicolumn{2}{c}{\textbf{Lisbon}} \\
    \cmidrule(lr){2-3} \cmidrule(lr){4-5}

    & Check-in & Population & Check-in & Population \\
    \midrule
    
    &$R^{2} \uparrow$ & $R^{2} \uparrow$ & $R^{2} \uparrow$ & $R^{2} \uparrow$ \\
    \midrule
    RegionDCL & 0.102 $\pm$ 0.033  &  0.408 $\pm$ 0.034  
    & 0.457 $\pm$ 0.056  & 0.586 $\pm$ 0.039
    \\
    UrbanCLIP & 0.136 $\pm$ 0.022  & 0.305 $\pm$ 0.008 
    & \underline{0.591 $\pm$ 0.012} & \underline{0.801 $\pm$ 0.004}
    \\
    CityFM & \underline{0.239 $\pm$ 0.039}  & \underline{0.412 $\pm$ 0.010} 
    & 0.195 $\pm$ 0.004 & 0.627 $\pm$ 0.003 
    \\
    \midrule 
    \textbf{\model} & \textbf{0.309 $\pm$ 0.027} & \textbf{0.581 $\pm$ 0.027} 
    & \textbf{0.831 $\pm$ 0.022} & \textbf{0.934 $\pm$ 0.008}
    \\
    \midrule 
    \textbf{Improvement} & \textbf{29.3\%} & \textbf{41.0\%} 
    &\textbf{40.6\%} & \textbf{16.6\%}
    \\

    \bottomrule
\end{tabular}
}
\vspace{-2mm}
\end{table}

Table~\ref{tab:model_adaptability_partial}  reports the results. We only report $R^2$ for conciseness, as the performance in MAE and RMSE resembles (Same below). 
\model\ outperforms all competitors consistently, with an improvement of at least 16.6\% in $R^2$ over the best baseline models. 
These results confirm \model's applicability across countries. Among the baseline models, CityFM performs better on Singapore, while UrbanCLIP is more suitable for Lisbon, underscoring the different urban characteristics of the two cities.

\subsection{Adaptability to Region Formations (Q3)}
\label{subsec:adaptability_of_grid_emb}



To evaluate the robustness and adaptability of our cell-based embeddings, we form regions of different sizes by recursively merging (with random region selection) the initial 180 regions (``\textbf{180r}'') of New York City with their randomly selected neighboring regions to form sets of 150 (``\textbf{150r}''), 120 (``\textbf{120r}''), and 90 (``\textbf{90r}'') regions.  
We repeat the experiments like above over each set of  regions. We omit results on the other cities as the patterns resemble (same below).


Note that \model\ only needs to learn the cell embeddings once, which can be reused to form  embeddings for the different sets of regions. 
In contrast, existing models, except for CityFM, require data reprocessing and model retraining for each set of regions. 

\begin{figure}[htbp]
    \centering
    \begin{subfigure}[b]{0.85\columnwidth}
        \centering
        \includegraphics[width=\columnwidth]{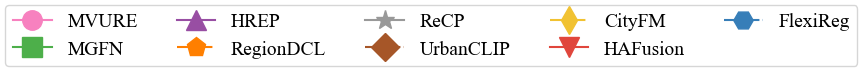}
    \end{subfigure}
    \begin{subfigure}[b]{0.47\columnwidth}
        \centering
        \includegraphics[width=0.9\columnwidth]{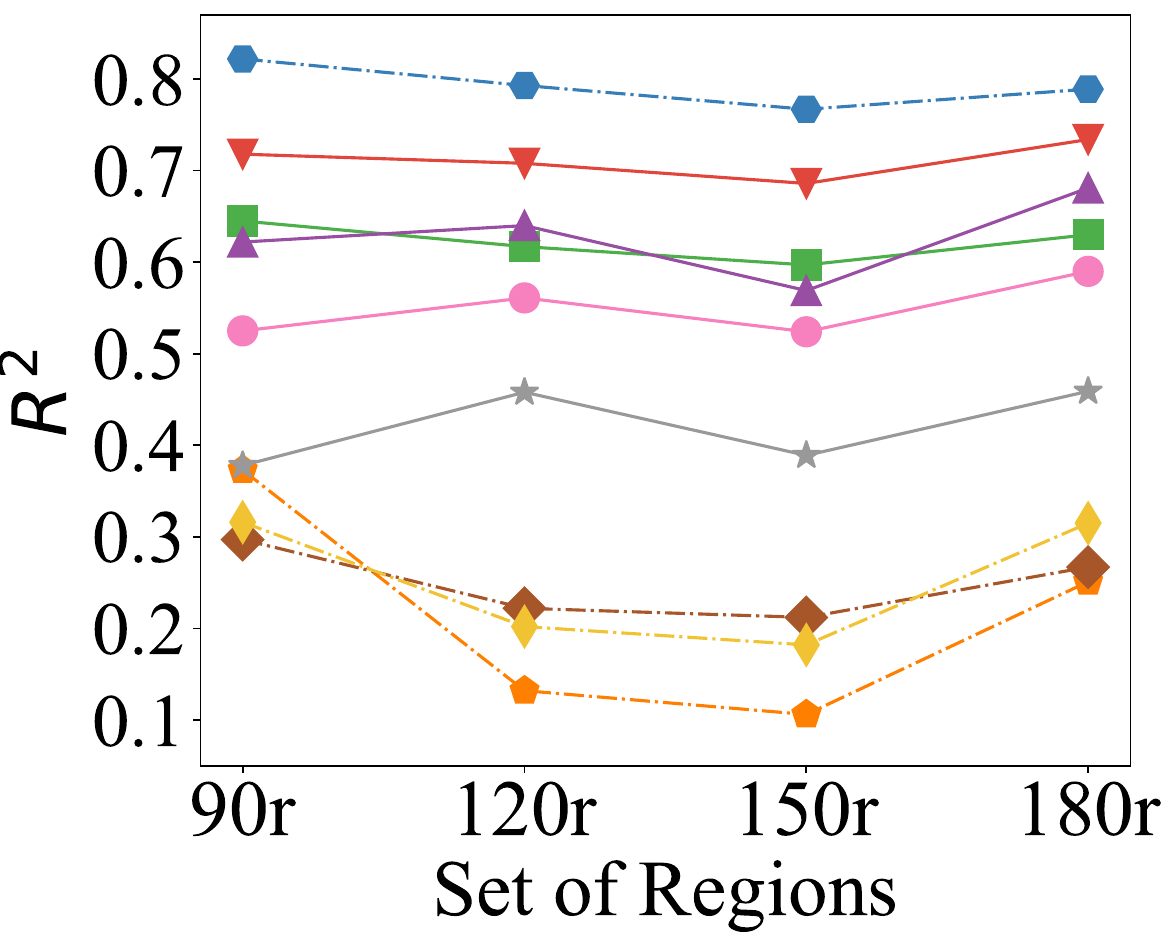}
        \vspace{-2mm}
        \caption{Crime}
    \end{subfigure}
    \begin{subfigure}[b]{0.47\columnwidth}
        \centering
        \includegraphics[width=0.9\columnwidth]{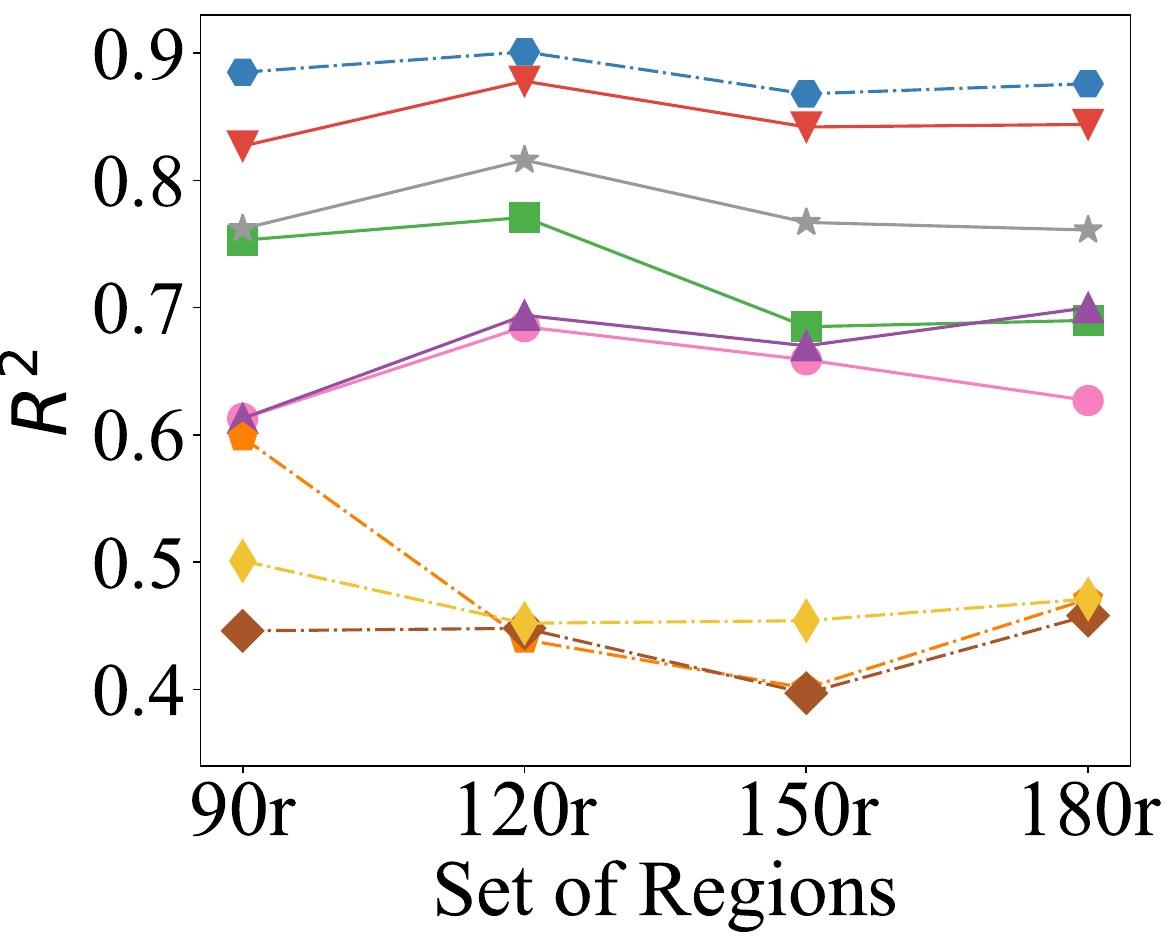}
        \vspace{-2mm}
        \caption{Check-in}
    \end{subfigure}

    \begin{subfigure}[b]{0.47\columnwidth}
        \centering
        \includegraphics[width=0.9\columnwidth]{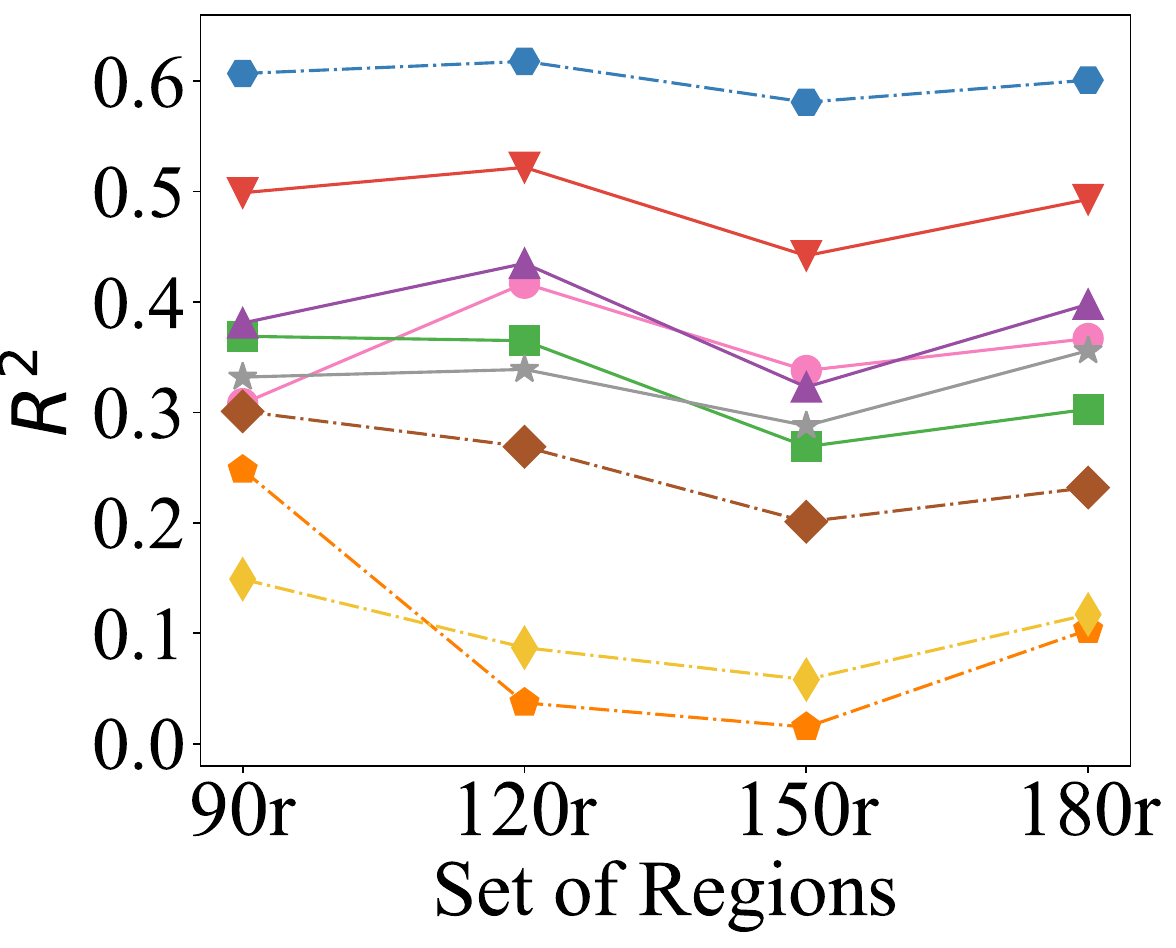}
        \vspace{-2mm}
        \caption{Service Call}
    \end{subfigure}
    \begin{subfigure}[b]{0.47\columnwidth}
        \centering
        \includegraphics[width=0.9\columnwidth]{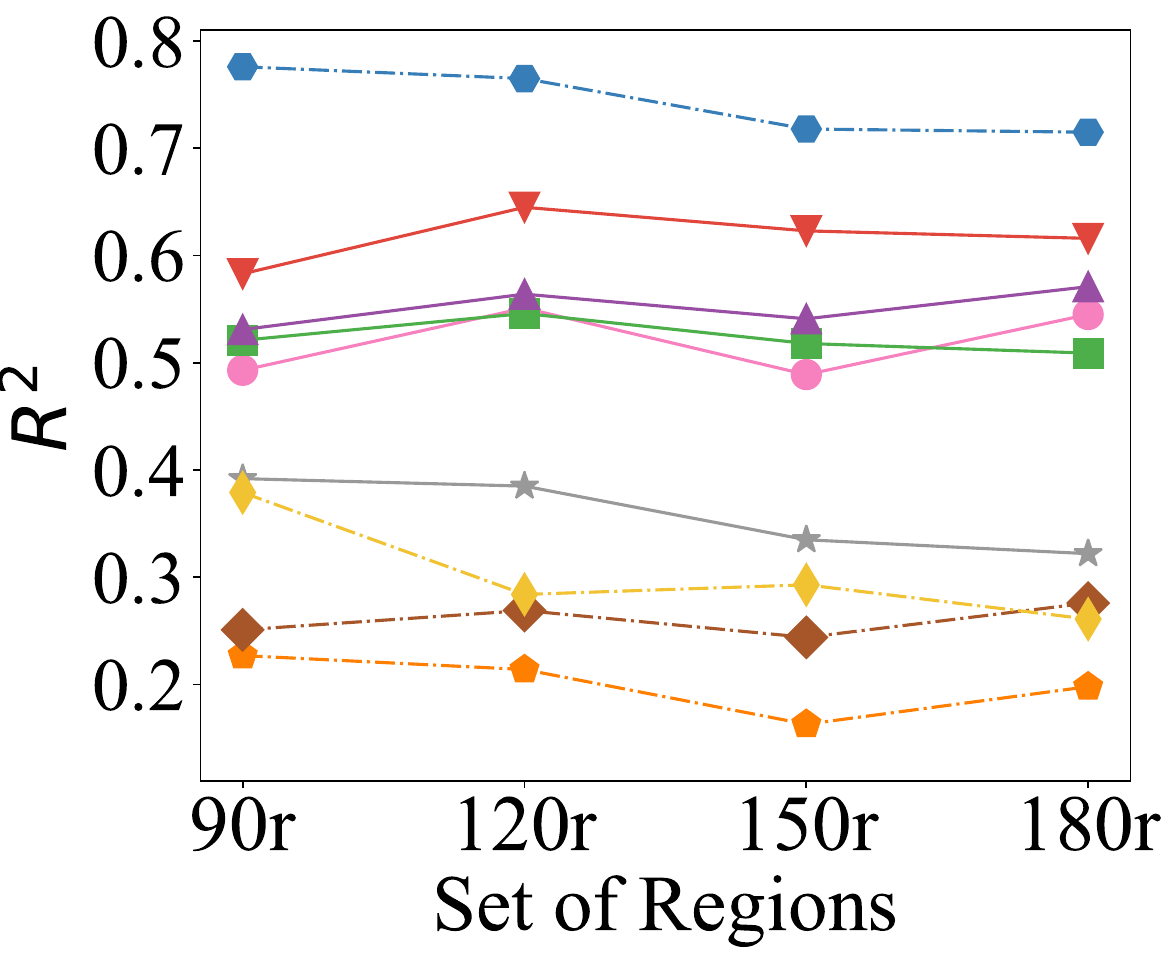}
        \vspace{-2mm}
        \caption{Population}
    \end{subfigure}
    \caption{Adaptability to regions of varying sizes (NYC).}
    \label{fig:grid_adaptability}
\end{figure}
 
As Fig.~\ref{fig:grid_adaptability} shows, \model\ again outperforms all competitors, with small performance variation (4\%) across different sets of regions. In comparison, the only existing model flexible to region formations, CityFM, struggles across all four sets of regions, outperformed by almost all the other baseline models that relearn the embeddings for each set of regions. 
These observations demonstrate the adaptability of our model and the effectiveness of \regionGen\ to accommodate different region formations.

\subsection{Ablation Study (Q4)}
\label{subsec:ablation_study}

We study the effectiveness of \model\ model components with the following variants:
(1)~\textbf{\model-w/o-PE} removes the prompt enhancer from the downstream task learning stage.
(2)~\textbf{\model-w/o-TAlign} replaces the text-region alignment module with a direct concatenation of text and region embeddings.
(3)~\textbf{\model-w/o-SVAlign} replaces the street view-region alignment module with the summation of all street view image embeddings in a region, followed by concatenating these with the region embeddings.
(4)~\textbf{\model-w/o-EC} removes the environment context-based contrastive learning and directly use ResNet to generate visual embeddings for each street view image.
(5) \textbf{\model-w/o-CNN} replaces the CNN branch with a GNN branch to process satellite images.
(6) \textbf{\model-w/o-Grid} directly learns region embeddings without using the grid cells.
(7) \textbf{\model-w/o-FS} uses all features (including textual descriptions and street view images) during cell embedding learning. 
(8) \textbf{\model-w/o-WS} replaces the weighted summation of cell embeddings with a direct summation.
(9) \textbf{\model-w/o-LT} uses the average of all token embeddings as the text embeddings instead of the last token embeddings.

\begin{figure}[htbp]
    \centering
    \begin{subfigure}[b]{\columnwidth}
        \centering
        \includegraphics[width=\columnwidth]{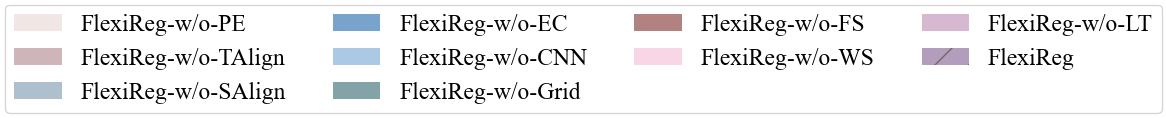}
    \end{subfigure}
    \begin{subfigure}[b]{0.48\columnwidth}
        \centering
        \includegraphics[width=\columnwidth]{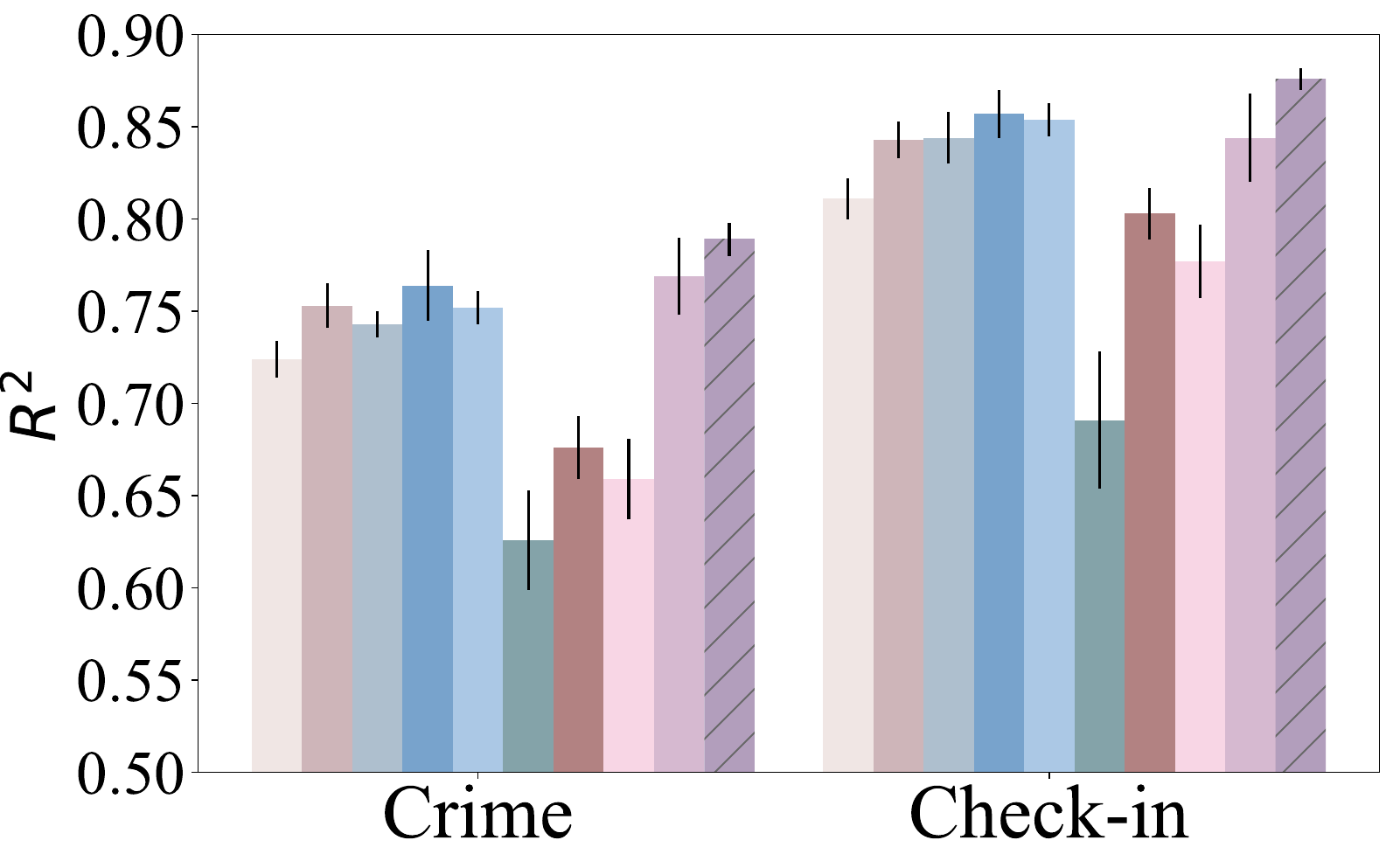}
    \end{subfigure}
    \begin{subfigure}[b]{0.48\columnwidth}
        \centering
        \includegraphics[width=\columnwidth]{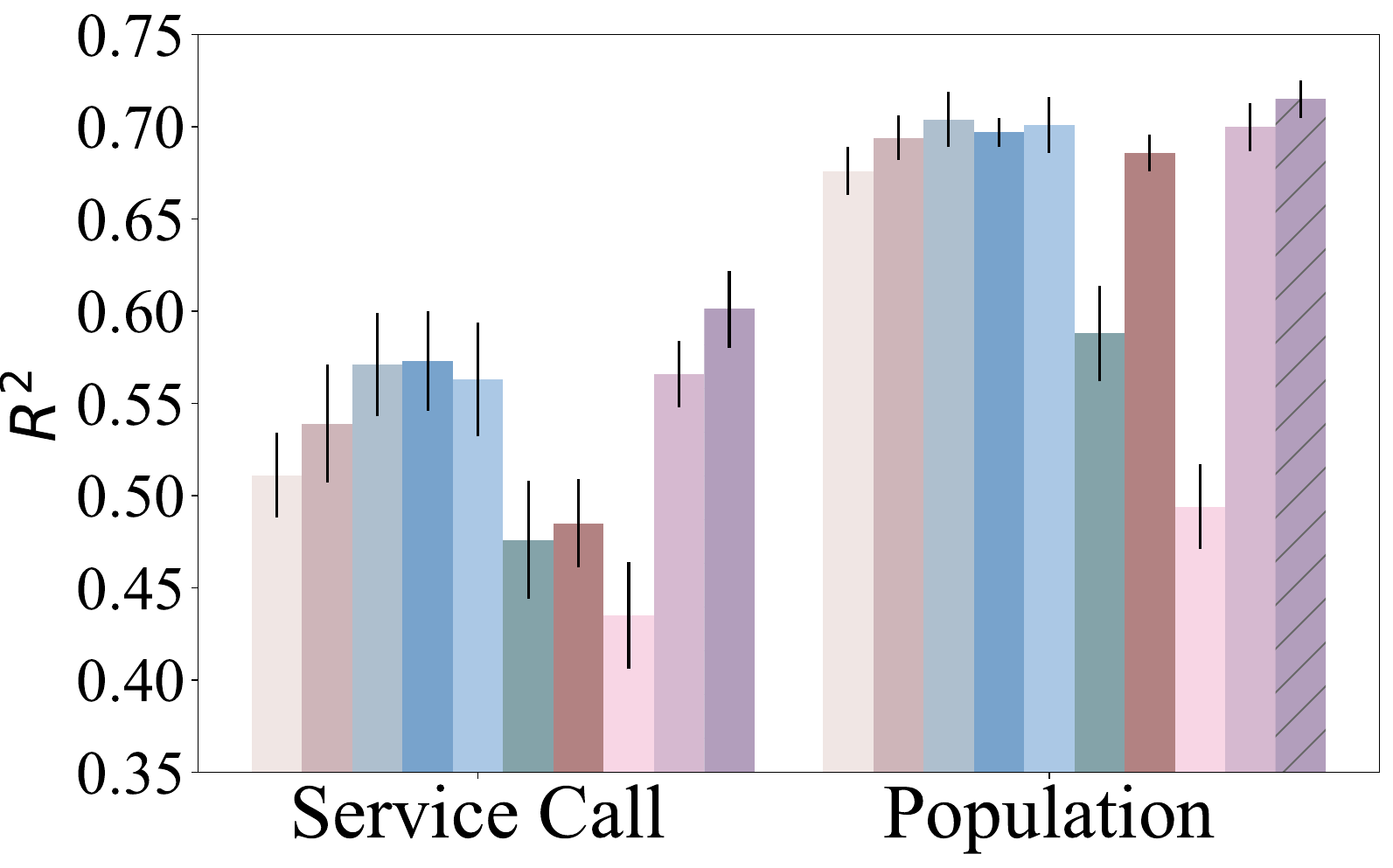}
    \end{subfigure}
    \caption{Ablation study results (NYC).} 
    \label{fig:ablation_model}
\end{figure}

We again repeat the experiments, and Fig.~\ref{fig:ablation_model} presents the results. 
As expected, \model\ consistently outperforms all  variants, highlighting the contribution of each model component to the overall effectiveness of \model. There are further observations:


(1)~\model-w-o-Grid is the worst across all tasks. This suggests that learning cell embeddings and then aggregating them into region embeddings contribute significantly to the overall model accuracy, as these steps enable the embeddings to better reflect local variations within regions. 
\model-w-o-WS also has low accuracy, implying that individual cell embeddings contribute differently to the region embeddings. A direct summation of cell embeddings could introduce noise and be even worse than not using cell embeddings at all (e.g., for  service call and population count prediction). 

(2)~\model-w-o-PE is also less accurate than \model, which emphasizes the need for the prompt enhancer module to tailor region embeddings to meet task-specific requirements.

(3) The low performance of \model-w-o-FS and \model-w-o-CNN highlights the importance of using appropriate methods to handle different features. Notably, \model-w-o-FS, which leverages all six features, performs worse than \model-w-o-PE that uses only four features. Simply incorporating more features does not necessarily enhance the embedding quality. 

\subsection{Impact of Input Features (Q4)}
\label{subsec:impact_of_diff_features}

To study the impact of input features, we exclude  each of the POI, land use, geographic neighbor, satellite image, textual description, and street view image features, forming six variants: \textbf{\model-w/o-P}, \textbf{\model-w/o-L}, \textbf{\model-w/o-N}, \textbf{\model-w/o-SI}, \textbf{\model-w/o-T}, and \textbf{\model-w/o-SV}, respectively. 

\begin{figure}[htbp]
    \centering
    \begin{subfigure}[b]{\columnwidth}
        \centering
        \includegraphics[width=0.85\columnwidth]{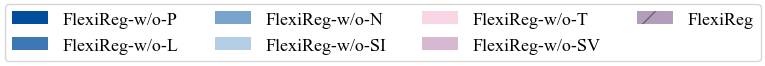}
    \end{subfigure}
    \begin{subfigure}[b]{0.48\columnwidth}
        \centering
        \includegraphics[width=\columnwidth]{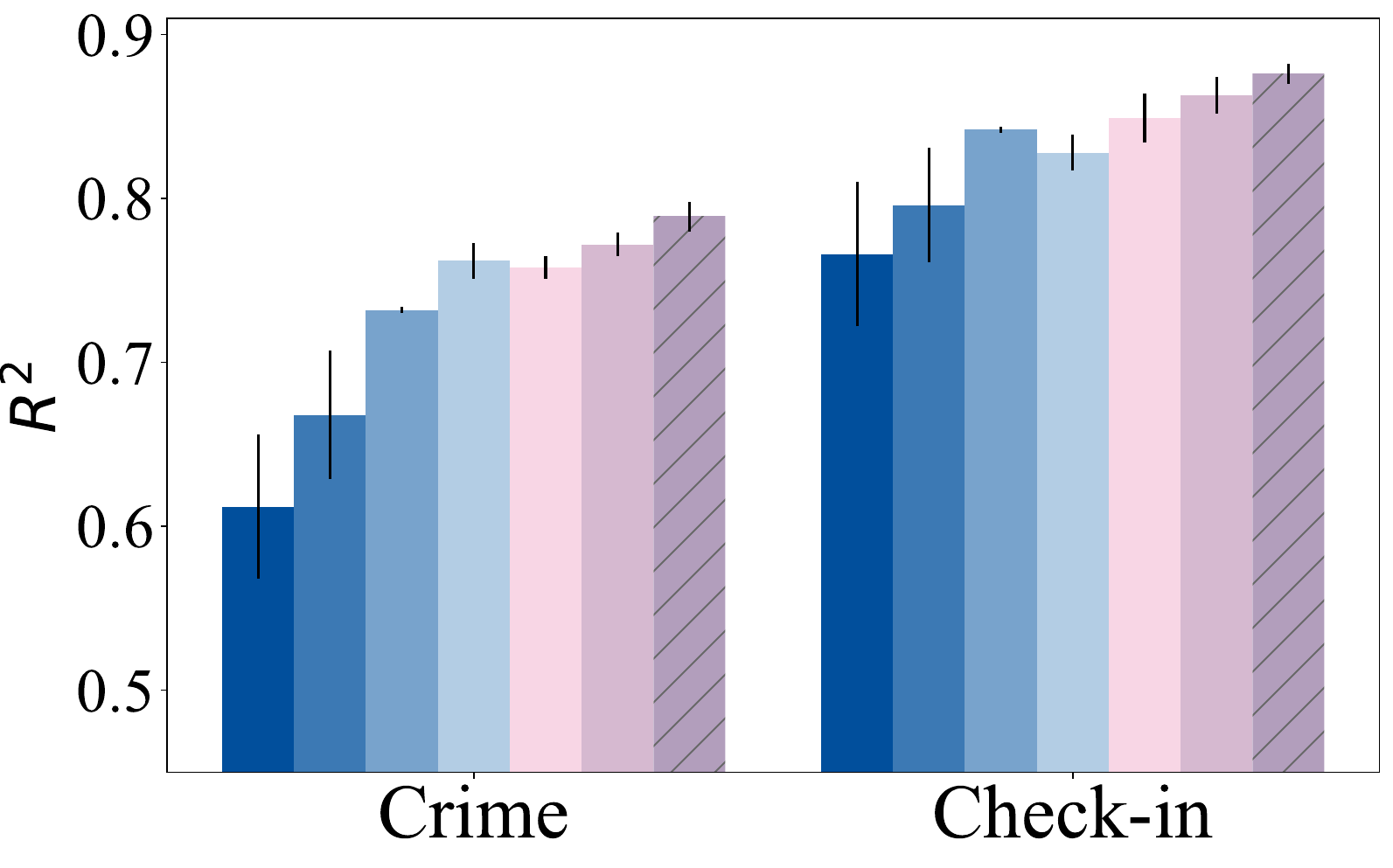}
    \end{subfigure}
    \begin{subfigure}[b]{0.48\columnwidth}
        \centering
        \includegraphics[width=\columnwidth]{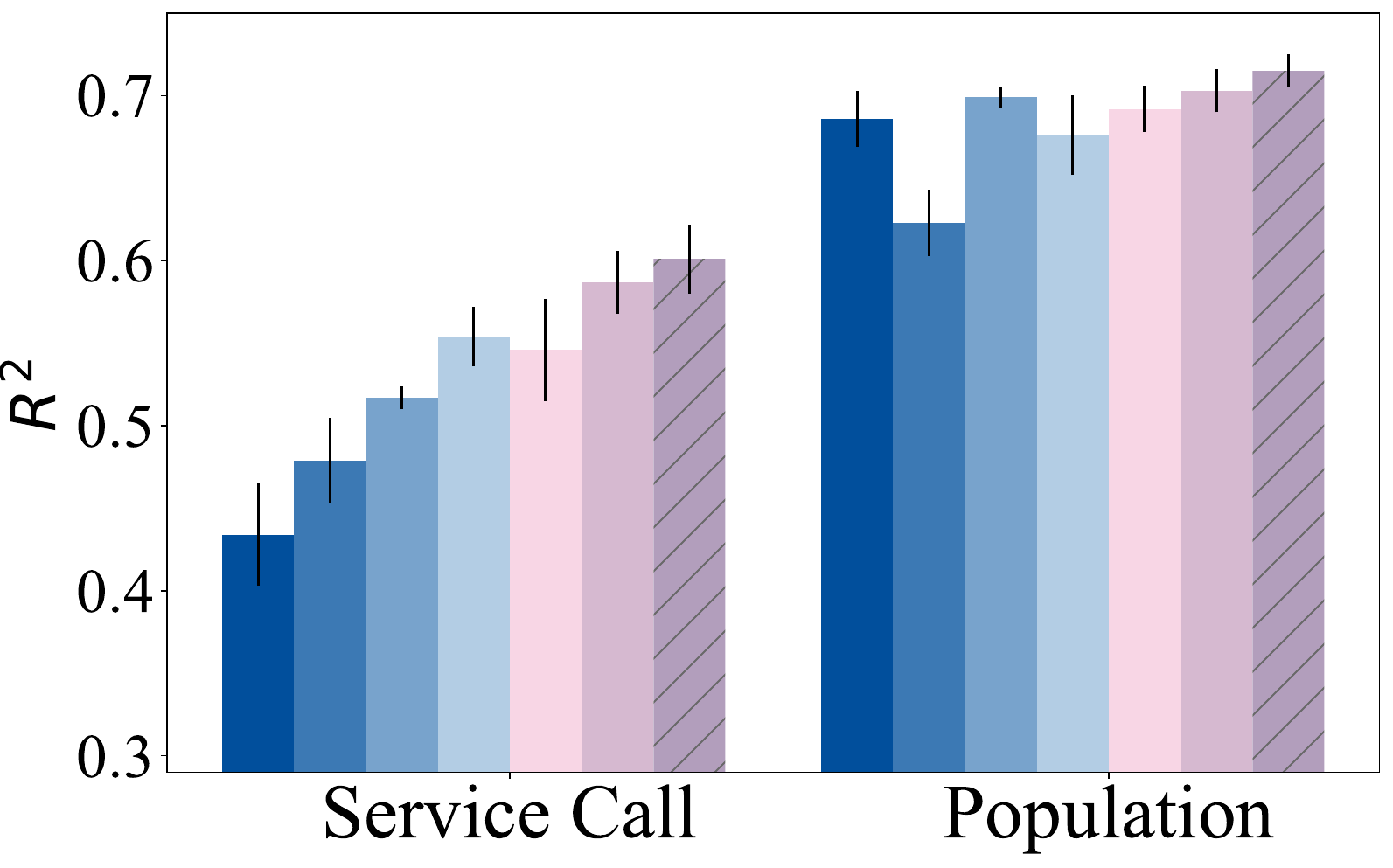}
    \end{subfigure}
    \caption{Impact of input features (NYC).}
    \label{fig:impact_of_views}
\end{figure}

We repeat the experiments and report results in Fig.~\ref{fig:impact_of_views}. The model variants are less accurate than \model\ that uses all input features, indicating the necessity of all these feature to achieve optimal embedding quality. 
\model-w/o-P has the lowest accuracy for crime, check-in, and service call count prediction. 
This is because POIs strongly correlate with human activities, which is a critical factor in these tasks. 
On the other hand, the land use feature contributes the most to the population count prediction task,  because it explicitly identifies areas for residential and other purposes, explaining for the low accuracy of \model-w/o-L for the task. 

Regarding the issue of street view image availability, \model-w/o-SV simulates scenarios where such imagery is unavailable for all regions. The results demonstrate that our model continues to achieve competitive performance by leveraging the remaining features and model components. 
Additionally, we perform a more fine-grained experiment to simulate partial availability, where 25\% of regions lack street view imagery, and we randomly sample street view images from their neighboring regions.
As shown in Table~\ref{tab:availability_steet_image}, the model’s performance remains largely unaffected, indicating its robustness to incomplete street view data.

\begin{table}[htbp] 
\caption{The availability of street view images ($R^2 \uparrow$ on NYC)}
\label{tab:availability_steet_image}
\begin{center}
\setlength{\tabcolsep}{4pt}
\begin{tabular}{l*{4}{c}}
    \toprule
     & Crime & Check-in & Service Call & Population \\
    \midrule
    
   100\% & \textbf{0.789} & \textbf{0.876} & \textbf{0.601} & \textbf{0.715} \\
    
    75\% & 0.785 & 0.870 & 0.595 & 0.711\\

    \bottomrule
\end{tabular}
\end{center}
\end{table}


\subsection{Applicability to Suburban Areas (Q5)}
\label{subsec:suburban_areas}

We also evaluate the applicability of our model to areas of different urban environments using regions in Staten Island, which is the largest borough of New York City by land area yet the least densely populated. This contrasts the area of the NYC dataset used above that covers Manhattan, which is the smallest borough by land area but the most densely populated. 

\begin{figure}[htbp]
  \centering
  \includegraphics[width=0.9\columnwidth]{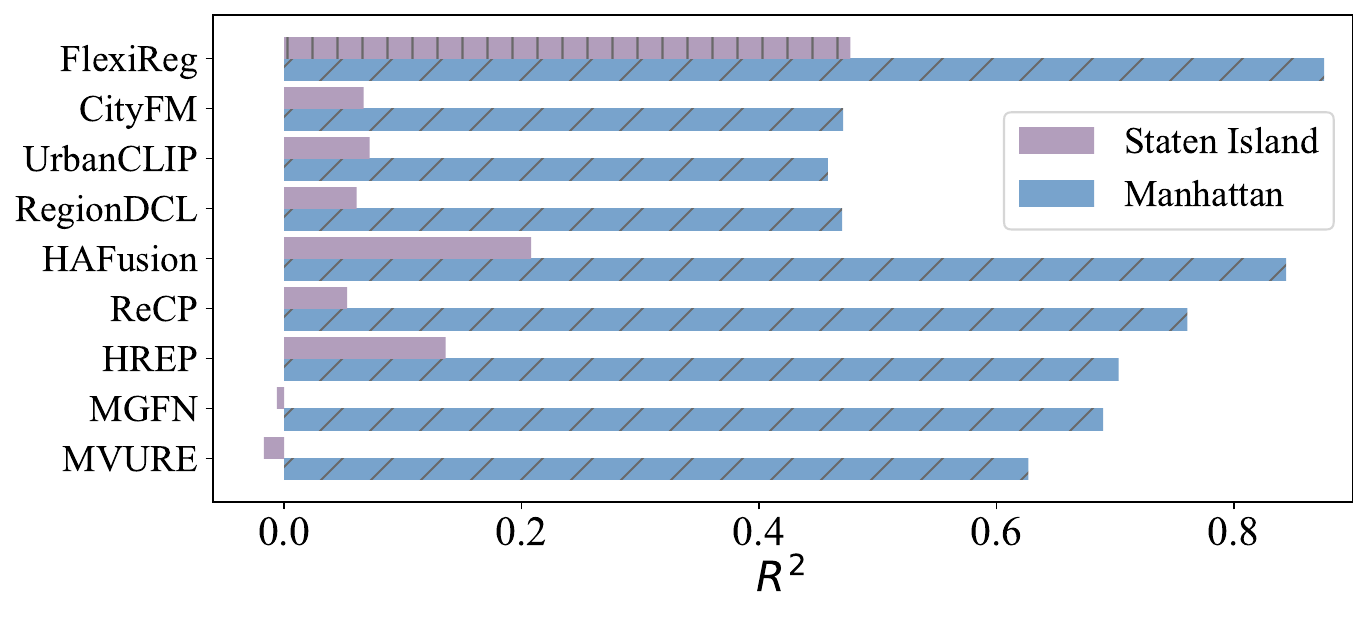}
  \vspace{-4mm}
  \caption{Model applicability to suburban areas (NYC).}
  \label{fig:population_density}
\end{figure}

We report in Fig.~\ref{fig:population_density} the $R^2$ results for the regions in these two boroughs, focusing on the check-in count prediction task for conciseness.
All models report lower accuracy (i.e., smaller $R^2$) on Staten Island than on Manhattan.  This is expected, as Staten Island is less densely populated with smaller variations in the urban features to help the models learn distinctive embeddings for different regions. The mobility data-based models suffer the most, as the mobility data become more sparse. 
\model\ again outperforms all competitors, now with an even larger performance gap. The model does not rely on mobility data, while its adaptive aggregation module enables it to adaptively fuse different input features of grid cells to accommodate areas of different urban characteristics.

\subsection{Impact of Grid Cell Design (Q6)}
\label{subsec:impact_grid_cell_design}
We use the following model variants to study the impact of shape and size of the grid cell:
(1)~\textbf{\model-Rect} uses a square grid instead of a hexagonal grid, where the edge length of each square is 200m.
(2)~\textbf{\model-LargeHex} uses a hexagonal grid with an edge length that is three times greater than the default length in \model.
(2) \textbf{\model-SmallHex} uses a hexagonal grid with an edge length that is one-third of the default length in \model.

As Table~\ref{tab:impact_of_grid_design} shows, \model\ produces the best (i.e., largest) $R^2$ scores across all four downstream tasks, confirming the effectiveness of our default grid cell design. 
\begin{table}[htbp] 
\caption{Impact of grid cell design ($R^2 \uparrow$ on NYC)}
\label{tab:impact_of_grid_design}
\begin{center}
\setlength{\tabcolsep}{4pt}
\resizebox{\columnwidth}{!}{
\begin{tabular}{l*{4}{c}}
    \toprule
     & Crime & Check-in & Service Call & Population \\
    \midrule
    
   \model-Rect & 0.741 & 0.836 & 0.557 & 0.699 \\
    
    \model-LargeHex & 0.587 & 0.784 & 0.425 & 0.233 \\

    \model-SmallHex & 0.709 & 0.803 & 0.586 & 0.710 \\

    \model & \textbf{0.798} & \textbf{0.876} & \textbf{0.601} & \textbf{0.715} \\
    \bottomrule
\end{tabular}
}
\end{center}
\end{table}

\model-Rect shares a similar number of cells with \model. Its lower $R^2$ scores indicate that the sqaure grid is less effective. We conjecture that  \model's better performance results from the symmetric structure of the hexagonal grid, where each cell has the same distance to all its neighbors, while this does not hold true when rectangular cells are used. Meanwhile, using hexagonal grid cells could better fit regions with irregular boundaries.

\model-LargeHex has the worst accuracy -- it has only 89 cells while \model\ has 438. The larger cells can mix the distinctive features in a cell, missing local urban characteristics. When the cells become larger than the target regions, capturing the distinctive urban features within a region becomes even more challenging. 
In contrast, \model-SmallHex has 3,201 cells. Now each cell is too small, and urban features become sparse, which makes it difficult to learn meaningful embeddings. These smaller cells also incur more processing times (57, 208, and 1294 seconds for \model-SmallHex, \model, and \model-LargeHex, respectively). These findings ground our choice of using a hexagonal grid.

\subsection{Impact of Prompt Templates (Q6)}
\label{subsec:impact_of_prompt_template}


As Fig.~\ref{fig:textual_discription_example} shows, we use a simple prompt template that contains only a task instruction (i.e., the first sentence) which will be populated with information specific to each related cell (e.g., its address).

\begin{table}[htbp] 
\caption{Impact of prompt templates ($R^2 \uparrow$ on NYC)}
\label{tab:impact_of_prompt_template}
\begin{center}
\setlength{\tabcolsep}{4pt}
\resizebox{\columnwidth}{!}{
\begin{tabular}{l*{4}{c}}
    \toprule
     & Crime & Check-in & Service Call & Population \\
    \midrule
    
   \model-ReM & 0.786 & 0.872 & 0.598 & 0.709 \\
    
    \model-RePh & \textbf{0.793} & 0.869 & 0.600 & 0.710 \\

    \model-RePos & 0.788 & 0.872 & \textbf{0.603} & 0.712 \\

    \model & 0.789 & \textbf{0.876} & 0.601 & \textbf{0.715} \\
    \bottomrule
\end{tabular}
}
\end{center}
\end{table}

We study the impact of prompt templates with three variants: (1)~\textbf{\model-ReM} removes the task instruction; (2)~\textbf{\model-RePh} rephrases the task instruction with ChatGPT; (3)~\textbf{\model-RePos} moves the task instruction to the end.

We repeat the embedding learning and downstream task prediction tasks as above. As Table~\ref{tab:impact_of_prompt_template} shows, the model variants achieve similar accuracy to \model, indicating that our model's effectiveness does not depend on a specific prompt template and is robust to prompt engineering. 
This robustness stems from our use of last-layer token embeddings from the LLM as textual embeddings, avoiding reliance on the generated text. Additionally, our proposed T-RAlign module effectively extracts task-relevant information from these embeddings.

\subsection{Impact of Model Parameter Values (Q6)}
\label{subsec:impact_of_parameter_learning}

We study model sensitivity to four key hyper-parameters: the dimensionality of the region embeddings ($d$), the dimensionality of the text-region embedding ($d_{text}$),  the number of street view images used per region in SV-RAlign ($\#SV$), and the pre-trained CNN model used in the CNN branch ($CNN_{PT}$). By default, we use the NYC dataset and report $R^2$ for conciseness in this subsection. 

\begin{figure}[htbp]
    \centering
    \begin{subfigure}[b]{\columnwidth}
        \centering
        \includegraphics[width=0.75\columnwidth]{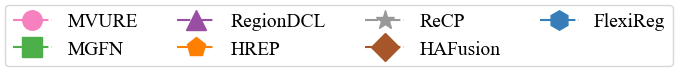}
    \end{subfigure}
    
    \begin{subfigure}[b]{0.47\columnwidth}
        \centering
        \includegraphics[width=0.9\columnwidth]{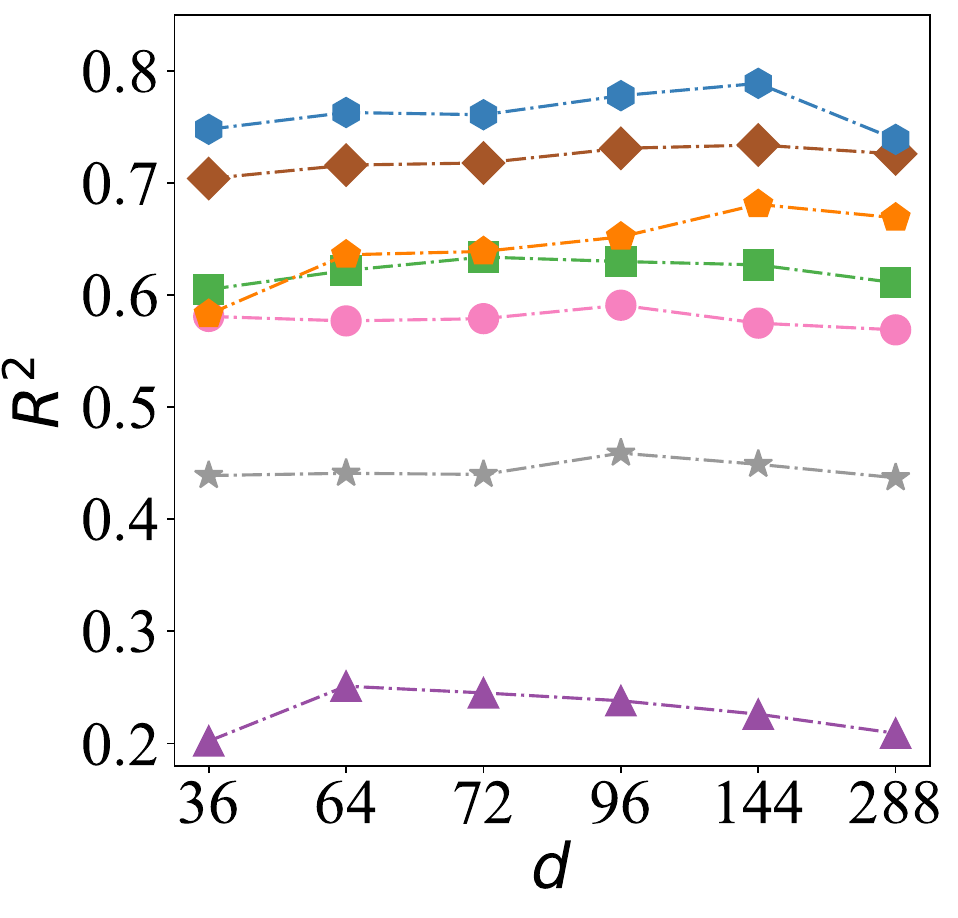}
        \vspace{-2mm}
        \caption{Crime}
    \end{subfigure}
    \begin{subfigure}[b]{0.47\columnwidth}
        \centering
        \includegraphics[width=0.9\columnwidth]{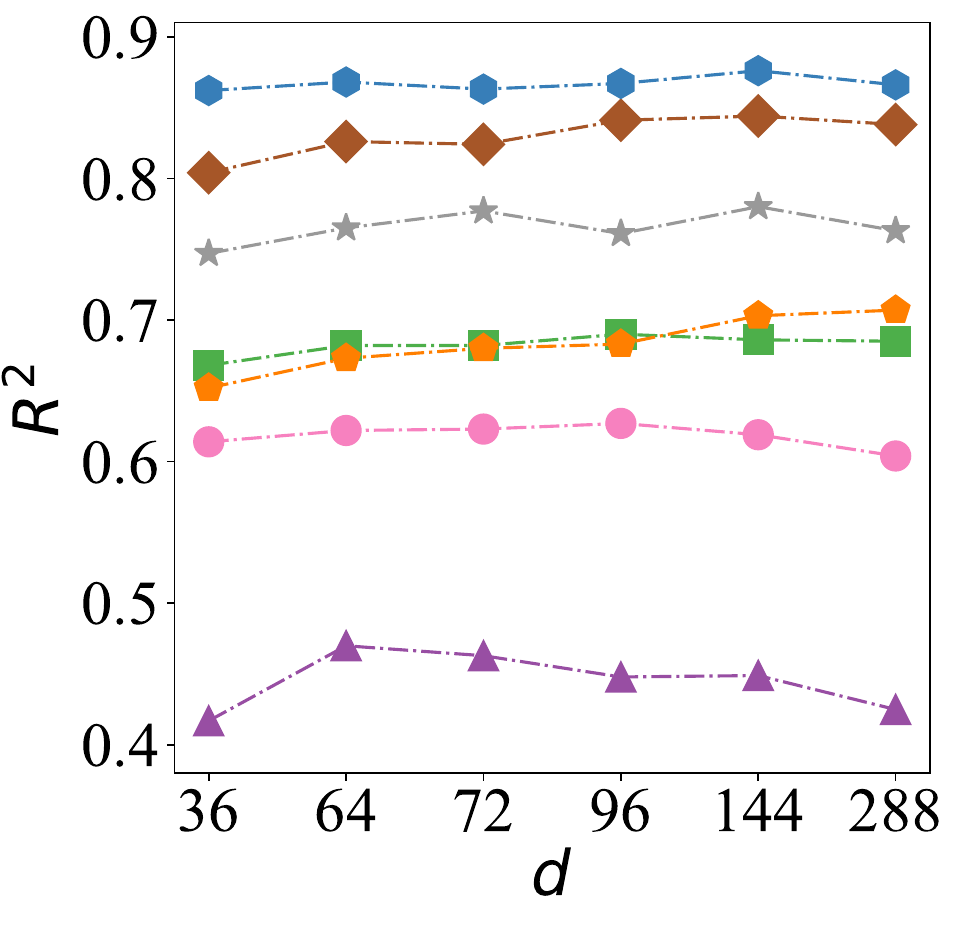}
        \vspace{-2mm}
        \caption{Check-in}
    \end{subfigure}

    \begin{subfigure}[b]{0.47\columnwidth}
        \centering
        \includegraphics[width=0.9\columnwidth]{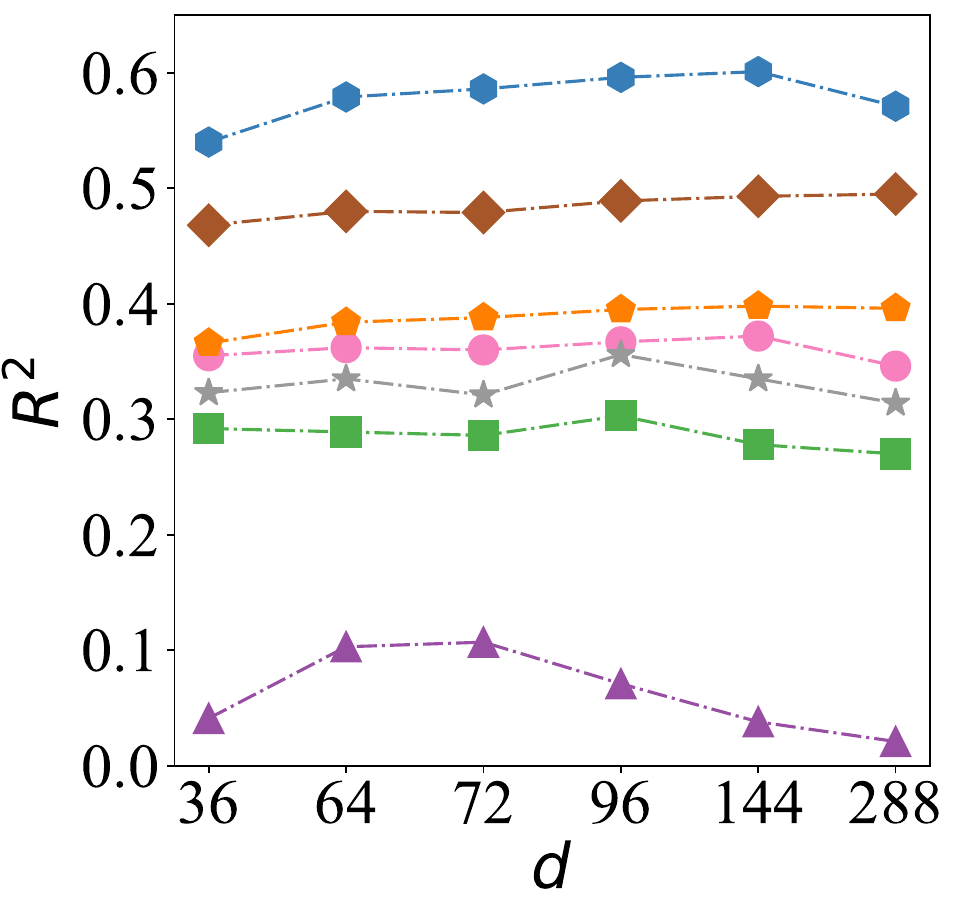}
        \vspace{-2mm}
        \caption{Service Call}
    \end{subfigure}
    \begin{subfigure}[b]{0.47\columnwidth}
        \centering
        \includegraphics[width=0.9\columnwidth]{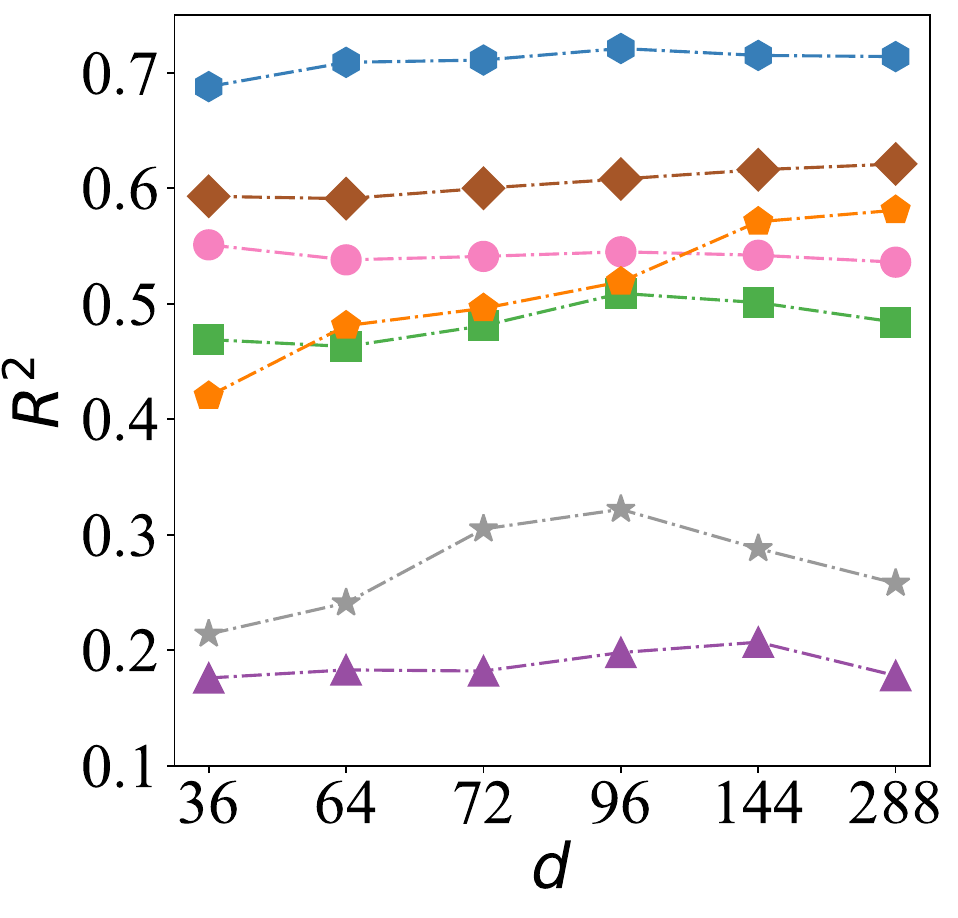}
        \vspace{-2mm}
        \caption{Population}
    \end{subfigure}
    \caption{Impact of $d$ (NYC).}
    \label{fig:impact_of_region_dim}
\end{figure}

\subsubsection{Impact of the region embedding dimensionality $d$.} 
\label{subsubsec:impact_of_region_dim}
We start by varying $d$ from 36 to 288. We test all models except for UrbanCLIP and CityFM whose $d$ cannot be easily varied as discussed in Section~\ref{subsec:exp_settings}. 
The learned region embeddings are used for the four downstream prediction tasks as before. 

Fig.~\ref{fig:impact_of_region_dim} shows that \model\ consistently outperforms all competitors across all tested values of $d$. Notably, the lowest $R^2$ value of \model\ on a task, regardless of the value of $d$, is at least as large as that of the maximum $R^2$ value that any baseline model can achieve on the same task. This emphasizes the robustness of \model\ over the embedding dimensionality. 

The optimal performance of different models is observed at varying  $d$  values. For our model, \model, the best performance across four downstream tasks is when  $d$  is between 96 and 144. We notice a drop in $R^2$ when $d$ becomes even larger, which is likely due to overfitting. Therefore, we default at $d$ = 144.
For the baseline models, the optimal  $d$  values align with the recommendations in their respective  papers. For example, HAFusion also reports  its best  performance at  $d$ = 144. We thus have used these values by default in the experiments above. 

\begin{table}[htbp] 
\caption{Impact of $d_{text}$ ($R^2 \uparrow$ on NYC)}
\label{tab:para_textEmb_dim}
\begin{center}
\setlength{\tabcolsep}{6pt}
\begin{tabular}{l*{5}{c}}
    \toprule
    $d_{text}$& 36 & 72 & 144 & 288 & 576\\
    \midrule
    
   Crime & 0.782 & 0.785 & \textbf{0.789} & 0.787 & 0.787 \\
    
    Check-in & 0.873 & 0.874 & \textbf{0.876} & 0.872 & 0.873 \\

    Service Call & 0.587 & 0.596 & 0.601 & \textbf{0.605} & 0.591 \\

    Population & 0.703 & 0.710 & \textbf{0.715} & 0.713 & 0.707 \\
    \bottomrule
\end{tabular}
\end{center}
\end{table}

\subsubsection{Impact of the text-region embedding dimensionality $d_{text}$.} 
The dimensionality of the text-region embeddings, $d_{text}$, impacts the expressiveness of the textual features in \model. We vary $d_{text}$ from 36 to 576 to study this impact. 
As shown in Table~\ref{tab:para_textEmb_dim}, \model' $R^2$ score improves as $d_{text}$ increases at start and then declines when $d_{text}$ exceeds 144 or 288.
Higher dimensionality enhances the expressiveness of the embeddings, enabling a better representation of the complex relationships between textual features and region embeddings. However, when the  dimensionality becomes too high, it may become redundant when the feature diversity is limited. This  also increases the risk of overfitting, where the embeddings become overly specific to the training data and lose generalizability to unseen data. Based on these observations, we have set $d_{text}$ to 144 by default.

\begin{table}[htbp] 
\caption{Impact of $\#SV$ ($R^2\uparrow$ on NYC)}
\label{tab:para_NO_SVimg}
\begin{center}
\setlength{\tabcolsep}{6pt}
\begin{tabular}{l*{5}{c}}
    \toprule
    $\#SV$& 16 & 32 & 64 & 128 & 256\\
     \midrule
    
   Crime & 0.787 & \textbf{0.789} & \textbf{0.789} & 0.785 & 0.781 \\
    
    Check-in & 0.872 & 0.875 & \textbf{0.876} & 0.870 & 0.864 \\

    Service Call & 0.587 & 0.597 & 0.601 & 0.598 & \textbf{0.603} \\

    Population & 0.707 & 0.712 & \textbf{0.715} & 0.713 & 0.708 \\
    \bottomrule
\end{tabular}
\end{center}
\end{table}

\subsubsection{Impact of the number of street view images  per region used in SV-RAlign, $\#SV$.} 
The number of street view images per region used in SV-RAlign reflects the richness of ground-level visual patterns captured. We vary $\#SV$ from 16 to 256 to evaluate its impact on the performance of \model. 
As Table~\ref{tab:para_NO_SVimg} shows, \model\ achieves the best performance with different values of $\#SV$ on different downstream tasks. The variation in $R^2$ for different $\#SV$ values on the same task is marginal, against verifying the robustness of our model. 
A larger $\#SV$ value, i.e., using more street view images, allows the model to capture more comprehensive visual patterns. It takes more time to process the images, and it could introduce noisy and conflicting patterns. 
We have set $\#SV$ as 64 by default to balance between learning effectiveness and efficiency.

\subsubsection{Impact of the pre-trained CNN model $CNN_{PT}$ used in the CNN branch.} 
The pre-trained CNN model impacts the capability of extracting features from satellite images. We explore  four different pre-trained CNN models (AlexNet~\cite{alexnet}, EfficientNet~\cite{efficientnet}, ResNet-18~\cite{resnet}, and ResNet-34~\cite{resnet}) for our model. For a clearer comparison over the impact of the use of these CNN models, we remove the prompt enhancer module from \model\ and focus solely on the multi-modal grid cell embedding learning module.

The results of Table~\ref{tab:para_used_CNN} show  that ResNet-18 yields the best accuracy. Compared to AlexNet and EfficientNet, the ResNet architecture uses residual connections which enable more effective extraction of both low-level and high-level features from satellite images. ResNet-18 has a relatively shallow architecture compared to ResNet-34, which makes it less prone to overfitting when dealing with less complex spatial patterns. Based on these results, ResNet-18 is selected as the default pre-trained CNN model for our model.

\begin{table}[htbp] 
\caption{Impact of $CNN_{PT}$ ($R^2 \uparrow$ on NYC)}
\label{tab:para_used_CNN}
\begin{center}
\setlength{\tabcolsep}{3pt}
\begin{tabular}{l*{4}{c}}
    \toprule
    $CNN_{PT}$ & AlexNet & EfficientNet & ResNet-18 & ResNet-34 \\
    \midrule
    
   Crime & 0.685 & 0.691 & \textbf{0.724} & 0.679 \\
    
    Check-in & 0.782 & 0.798 & \textbf{0.811} & 0.787 \\

    Service Call & 0.439 & 0.454 & \textbf{0.511} & 0.443 \\

    Population & 0.660 & 0.649 & \textbf{0.676} & 0.667 \\
    \bottomrule
\end{tabular}
\end{center}
\end{table}


\section{Related Work}
Existing studies have achieved success from various perspectives. We categorize them into different groups from three perspectives, as presented in Table~\ref{tab:related_work}.

Human mobility data have been extensively utilized in existing studies. Some studies~\cite{ZE-Mob, CDAE, MGFN, RAW} rely solely on human mobility data, limiting their ability to capture more comprehensive urban region characteristics from different perspectives.
In contrast, most studies~\cite{HAFusion, HDGE, MP-VN, CGAL, ReMVC, DLCL, MVURE, ReCP, CGAP, MuseCL, MGRL4RE} integrate human mobility data with other types of features. Among these studies, some~\cite{HDGE, MP-VN, CGAL, ReMVC, MuseCL} employ MLP-based or CNN-based view encoders to capture information from each type of features. 
These studies focus on individual regions and consider less  the correlations between them. Several studies~\cite{MVURE, DLCL, HREP, ReCP, CGAP, MGRL4RE} represent regions and  their relationships using a graph structure and apply GNNs to learn the correlations between the regions. 
Recent studies further improve the integration of information from multiple views. 
HAFusion~\cite{HAFusion} proposes an attention-based fusion module to capture the high-order correlation between regions,  while ReCP~\cite{ReCP} replaces traditional fusion modules with a contrastive learning framework. This framework optimizes two objectives: maximizing mutual information between different views and minimizing conditional entropy across them. For these studies, the core issue is that human mobility data are only available for certain regions, limiting their applicability to different regions.

Recently, studies have shifted focus to learning region embeddings from publicly accessible data, such as POIs and satellite images.
HGI~\cite{HGI} mainly leverages POI features and designs a hierarchical graph structure
to build connectivity at both the POI and region levels. Then, it applies a GNN to learn POI embeddings and  aggregates the embeddings to the region level.
Urban2Vec~\cite{urban2vec} and M3G~\cite{m3g} utilize street view images and POI textual descriptions to learn region embeddings through a triplet loss framework. 
RegionDCL~\cite{RegionDCL} leverages building footprints by partitioning them into non-overlapping groups. It then applies contrastive learning at both the building-group and region levels to generate building-group embeddings, which are aggregated to generate the region embedding.
PG-SimCLR~\cite{PG-SimCLR} trains an image encoder on satellite images using contrastive learning based on spatial proximity and POI category distributions. The visual representation of satellite images is used as  the corresponding region embedding. 
MMGR~\cite{MMGR} employs two encoders: one for POI categories and the other for satellite images, generating POI and visual embeddings, respectively. It fuses these embeddings using contrastive learning to produce region embeddings. 
UrbanCLIP~\cite{urbanclip} utilizes a vision language model to generate detailed textual descriptions for satellite images corresponding to regions. The model then learns visual region embeddings by processing image-text pairs via contrastive learning.
GeoVectors~\cite{GeoVectors} and CityFM~\cite{cityFM} generate embeddings for various geospatial entities from OpenStreetMap, which are then aggregated to produce region embeddings. GeoVectors utilizes random walks to learn representations of entities’ locations and employs FastText~\cite{FastText} to learn representations of entities' textual annotations, combining both to form the final entity embeddings.
CityFM proposes a contrastive learning framework with three  objectives: a mutual information-based text-to-text objective, a vision-language objective, and a road-based context-to-context objective.
However, these models fall behind mobility-based models on downstream task, as they fail to capture underlying spatial correlations between different features. 

These existing studies primarily focus on the general region embedding learning stage, while paying limited attention to the downstream task learning phase. Prompting learning provides a promising solution for downstream task-based learning, which can guide the general region embeddings to adapt to specific tasks. This approach has already achieved significant success in fields such as natural language processing~\cite{nlp1, nlp2, nlp3} and computer vision~\cite{CV1, CV2, CV3}. 
Recently, prompt learning has also been introduced into the field of urban computing~\cite{zhang2023promptst, yuan2024unist}.  
However, the urban tasks targeted in these works differ in focus from ours.

In the context of region embedding learning, only one prior attempt exists: HREP [62], which generates prompt embeddings randomly, without conditioning on any input features. While the prompt embeddings may be correlated with the downstream tasks, they fail to capture the correlation between the region features and downstream tasks.

\textbf{Discussion.}
Existing high-performing models rely heavily on human mobility data, while accessible features and their relevance to downstream tasks have been underexplored. We propose a novel, multimodal grid cell embedding learning  module and an environment context-based contrastive learning approach to capture distinctive environmental characteristics and spatial correlations between different types of features.  Besides, we introduce a prompt enhancer module to extract and integrate task-specific information into generic region embeddings to tailor for diverse tasks.

\section{Conclusion}
\label{sec:conclusion}

We proposed a novel urban region representation learning model named \model\ towards generating flexible representations to accommodate the needs of different downstream tasks with different region formations.
\model\ only requires public accessible data. It learns fine-grained grid cells independent of the region partitions used by downstream tasks, achieving region formation flexibility. \model\ comes with a multi-modal grid cell embedding learning module and an adaptive region embedding generation module to learn cell and region embeddings, respectively. It further incorporates a prompt enhancer module to extract task-specific information and integrate such information into region embeddings to achieve downstream task flexibility. Extensive experiments on real-world dataset from cities in different countries show that \model\ significantly outperforms all SOTA models across different downstream tasks in diverse geographic regions.


\bibliographystyle{ACM-Reference-Format}
\bibliography{ref}



\end{document}